\documentclass[10pt,twocolumn,letterpaper]{article}

\usepackage{wacv}
\usepackage{times}
\usepackage{epsfig}
\usepackage{graphicx}
\usepackage{amsmath}
\usepackage{amssymb}

\usepackage{colortbl}
\usepackage{tikz,pgfplots}

\definecolor{rowgray}{rgb}{0.9275,0.9275,0.9275}
\definecolor{note}{rgb}{0.1,0.1,1}

\usepackage[pagebackref=true,breaklinks=true,letterpaper=true,colorlinks,bookmarks=false]{hyperref}

\wacvfinalcopy 

\def\wacvPaperID{49} 

\ifwacvfinal\fi
\begin{document}

\title{Video Object Segmentation-based Visual Servo Control \\ and Object Depth Estimation on a Mobile Robot}

\author{Brent A. Griffin \hspace{1cm} Victoria Florence \hspace{1cm} Jason J. Corso \\
	University of Michigan\\
	{\tt\small \{griffb,vflorenc,jjcorso\}@umich.edu}}

\maketitle
\ifwacvfinal\thispagestyle{empty}\fi

\begin{abstract}
	
		To be useful in everyday environments, robots must be able to identify and locate real-world objects.
		In recent years, video object segmentation has made significant progress on densely separating such objects from background in real and challenging videos.
		Building off of this progress, this paper addresses the problem of identifying generic objects and locating them in 3D using a mobile robot with an RGB camera.
		We achieve this by, first, introducing a video object segmentation-based approach to visual servo control and active perception and, second, developing a new Hadamard-Broyden update formulation.
		Our segmentation-based methods are simple but effective, and our update formulation lets a robot quickly learn the relationship between actuators and visual features without any camera calibration.
		We validate our approach in experiments by learning a variety of actuator-camera configurations on a mobile HSR robot, which subsequently identifies, locates, and grasps objects from the YCB dataset and tracks people and other dynamic articulated objects in real-time.

	
\end{abstract}

\begin{figure}
	\centering
	\includegraphics[width=0.4625\textwidth]{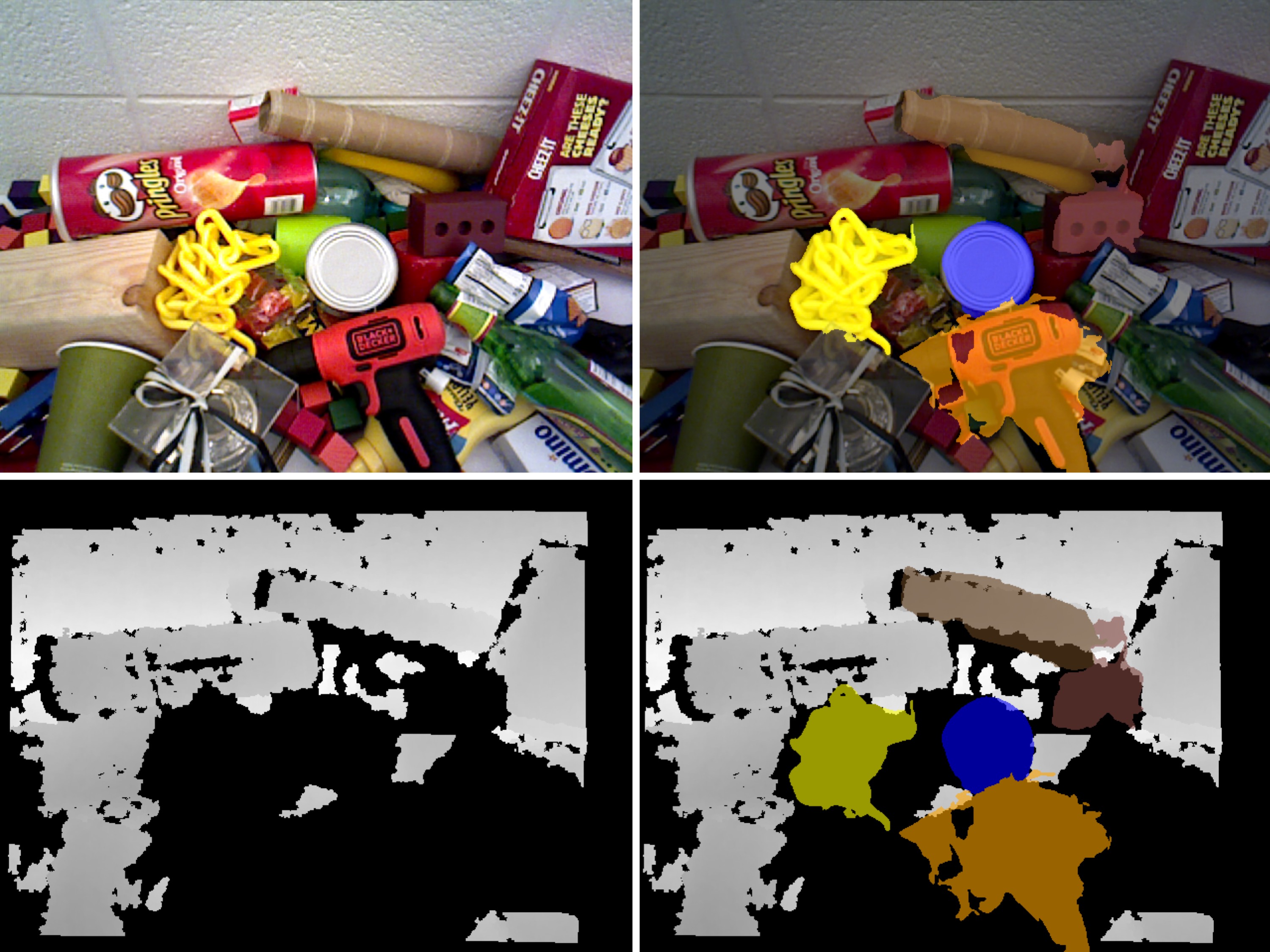}
	\caption{\textbf{RGBD View of Cluttered Scene}. Using an RGB image (top left), HSR identifies and segments five target objects (top right). However, the associated depth image is unreliable (bottom left) and provides depth data for only one target (bottom right).
	}
	\label{fig:rgbd}
\end{figure}
\begin{figure}
	\centering
	\includegraphics[width=0.4625\textwidth]{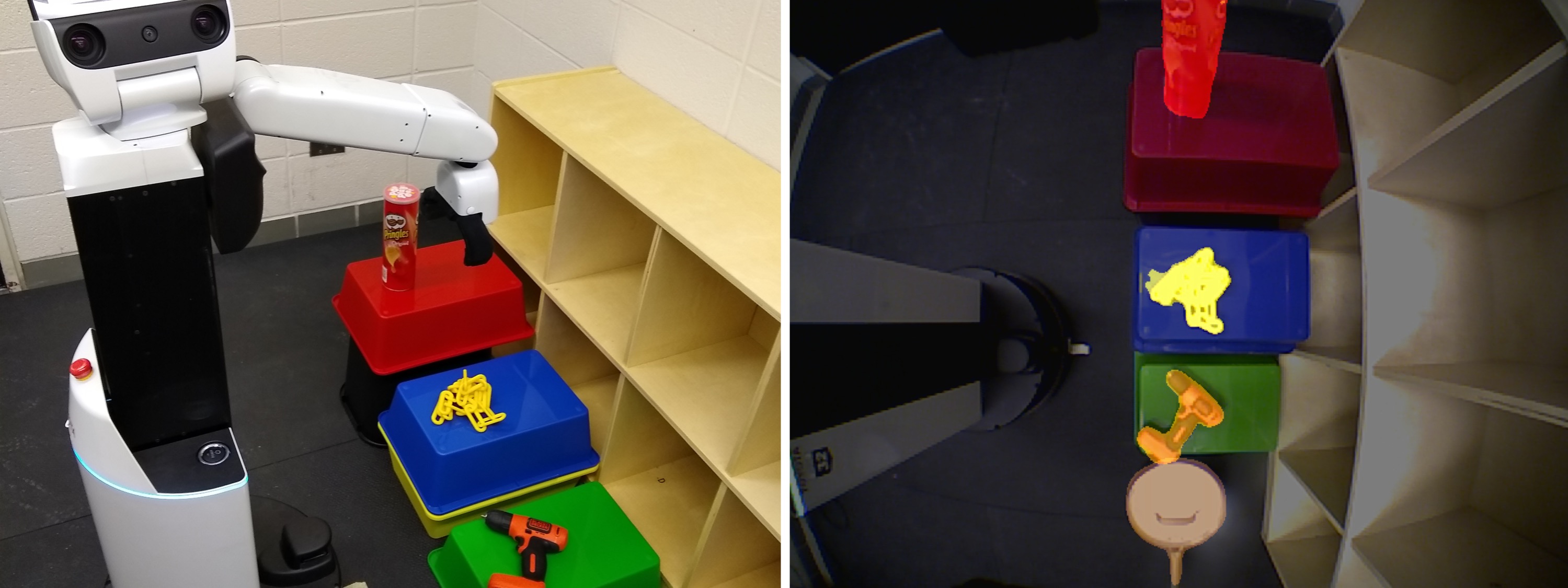}
	\caption{\textbf{Finding Objects in RGB}.
		With our approach, HSR segments, locates, and grasps objects using a single RGB camera.}
	\label{fig:stairs}
\end{figure}

\section{Introduction}
\label{sec:intro}

Visual servo control (VS), using visual data in the servo loop to control a robot, is a well-established field \cite{ChHu06,HuHaCo96}.
Using features from RGB images, VS has been used for positioning UAVs \cite{GuHaMa08,McJaCo17} and wheeled robots \cite{LuOrGi08,MaOrPr07}, manipulating objects \cite{JaFuNe97,WaLaSi10}, and even laparoscopic surgery \cite{WeArHi97}.
While this prior work attests to applicability of VS, generating robust visual features for VS in unstructured environments with generic objects (e.g., without fiducial markers) remains an open problem.


On the other hand, video object segmentation (VOS), the dense separation of objects in video from background, has made recent progress on real, unstructured videos.
This progress is due in part to the introduction of multiple benchmark datasets \cite{DAVIS, DAVIS17, YTVOS}, which evaluate VOS methods across many challenging categories, including moving cameras, occlusions, objects leaving view, scale variation, appearance change, edge ambiguity, multiple interacting objects, and dynamic background; these challenges frequently occur simultaneously.
However, despite all of VOS's contributions to video understanding, we are unaware of any work that utilizes VOS for control.

To this end, this paper develops a VOS-based framework to address the problem of visual servo control in unstructured environments.
We also use VOS to estimate depth without a 3D sensor (e.g., an RGBD camera in Figure~\ref{fig:rgbd} and \cite{FeLa19,WaEtAl19}).
Developing VOS-based features for control and depth estimation has many advantages.
First, VOS methods are robust across a variety of unstructured objects and backgrounds, making our framework general to many settings.
Second, many VOS methods operate on streaming images, making them ideal for tracking objects from a moving robot.
Third, ongoing work in active and interactive perception enables robots to automatically generate object-specific training data for VOS methods \cite{BrTiMeBuWa14,FlCoGr19,KrHeReFo11,MaFlMaTe18}. 
Finally, VOS remains a hotly studied area of video understanding, and future improvements in the accuracy and robustness of state-of-the-art segmentation methods will similarly improve our method.

The primary contribution of our paper is the development and experimental evaluation of video object segmentation-based visual servo control (VOS-VS).
We demonstrate the utility of VOS-VS on a mobile robot equipped with an RGB camera to identify and position itself relative to many challenging objects from HSR challenges and the YCB object dataset \cite{YCB}. 
To the best of our knowledge, this work is first use of video object segmentation for control.

A second contribution is our new Hadamard-Broyden update formulation, which outperforms the original Broyden update in experiments and enables a robot to learn the relationship between actuators and VOS-VS features online without any camera calibration.
Using our update, our robot learns to servo with seven unique configurations across seven actuators and two cameras.
To the best of our knowledge, this work is the first use of a Broyden update to directly estimate the \textit{pseudoinverse} feature Jacobian for visual servo control on a robot.

A final contribution is introducing two more VOS-based methods, VOS-DE and VOS-Grasp. 
VOS-DE combines segmentation features with Galileo's Square-cube law and active perception to estimate an object's depth, which, with VOS-VS, provides an object's 3D location.
VOS-Grasp uses segmentation features for grasping and grasp-error detection.
Thus, using our approach, robots can find and grasp objects using a single RGB camera (see Figure~\ref{fig:stairs}).

We provide source code and annotated YCB object training data at \url{https://github.com/griffbr/VOSVS}.





%











\section{Related Work}

%
%

\subsection{Video Object Segmentation}

Video object segmentation methods can be categorized as unsupervised, which usually rely on object motion \cite{NLC,GrCoWACV2019,KEY,FST,WeSz17}, or semi-supervised, which segment objects specified in user-annotated examples \cite{CINM,PML,GrCo19,PREMVOS,RGMP,OSMN}.
Of particular interest to the current work, semi-supervised methods learn the visual characteristics of a target object, which enables them to reliably segment dynamic \textit{or} static objects.
To generate our VOS-based features, we segment objects using One-Shot Video Object Segmentation (OSVOS) \cite{OSVOS}, 
which is state-of-the-art in VOS and has influenced other leading semi-supervised methods \cite{OSVOS-S,OnAVOS}.  

\subsection{Visual Servo Control}

In addition to the visual servo literature cited in Section~\ref{sec:intro}, this paper builds off of other methods for control design and feature selection.
For control design, a technique using a hybrid input of 3D Cartesian space and 2D image space is developed in \cite{MaChBo99}, with depth estimation provided externally.
As a step toward more natural image features, Canny edge detection-based planar contours of objects are used in \cite{ChMaCi00}.
When designing features, work in \cite{MaCoCh02} shows that  $z$-axis features should scale proportional to the optical depth of observed targets.
Finally, work in \cite{CoHu01} controls $z$-axis motions using the longest line connecting two feature points for rotation and the square root of the collective feature-point-polygon area for depth; 
this approach addresses the Chaumette Conundrum presented in \cite{Ch98} but also requires that all feature points remain in the image.
Notably, early VS methods require structured visual features (e.g., fiducial markers), while recent learning-based methods require manipulators with a fixed workspace \cite{AbEtAl19,JaEtAl19,ZuEtAl19}.

Taking advantage of recent progress in computer vision, this paper introduces robust segmentation-based image features for visual servoing that are generated from ordinary, real-world objects.
Furthermore, our features are rotation invariant, work when parts of an object are out of view or occluded, and do not require any particular object viewpoint or marking, making this work applicable to articulated and deformable objects (e.g., the yellow chain in Figures~\ref{fig:rgbd}-\ref{fig:stairs}).
Finally, our method enables visual servo control on a mobile manipulation platform, on which we also use segmentation-based features for depth estimation and grasping.

\subsection{Active Perception}

A critical asset for robot perception is taking actions to improve sensing and understanding of the environment, i.e., Active Perception (AP) \cite{Bajcsy_active_perc,Bajcsy2018_active_perc}.
Compared to structure from motion \cite{SFM_survey,KaGaBa19,LonguetHiggins198761}, which requires feature matching or scene flow to relate images, AP exploits knowledge of a robot's relative position to relate images and improve 3D reconstruction.
Furthermore, AP methods select new view locations explicitly to improve perception performance \cite{Eidenberger_IROS2010,SpRoCh17,zeng2018robotic_amazon_pick_place}.
In this work, we use active perception with VOS-based features to estimate an object's depth.
We complete our estimate during our robot's approach to an object, and, by tracking the estimate's convergence, we can collect more data if necessary.
Essentially, by using an RGB camera and kinematic information that is already available,  we estimate the 3D position of objects without any 3D sensors, including: LIDAR, which is cost prohibitive and color blind; RGBD cameras, which do not work in ambient sunlight among other conditions (see Figure~\ref{fig:rgbd}); and stereo cameras, which require calibration and feature matching.
Even when 3D sensors are available, RGB-based methods provide an indispensable backup for perception \cite{MiPhEtAl17,McBoEtAl17}.

%
%


%


\section{Robot Model and Perception Hardware}

\begin{figure}
	\centering
	\includegraphics[width=0.4625\textwidth]{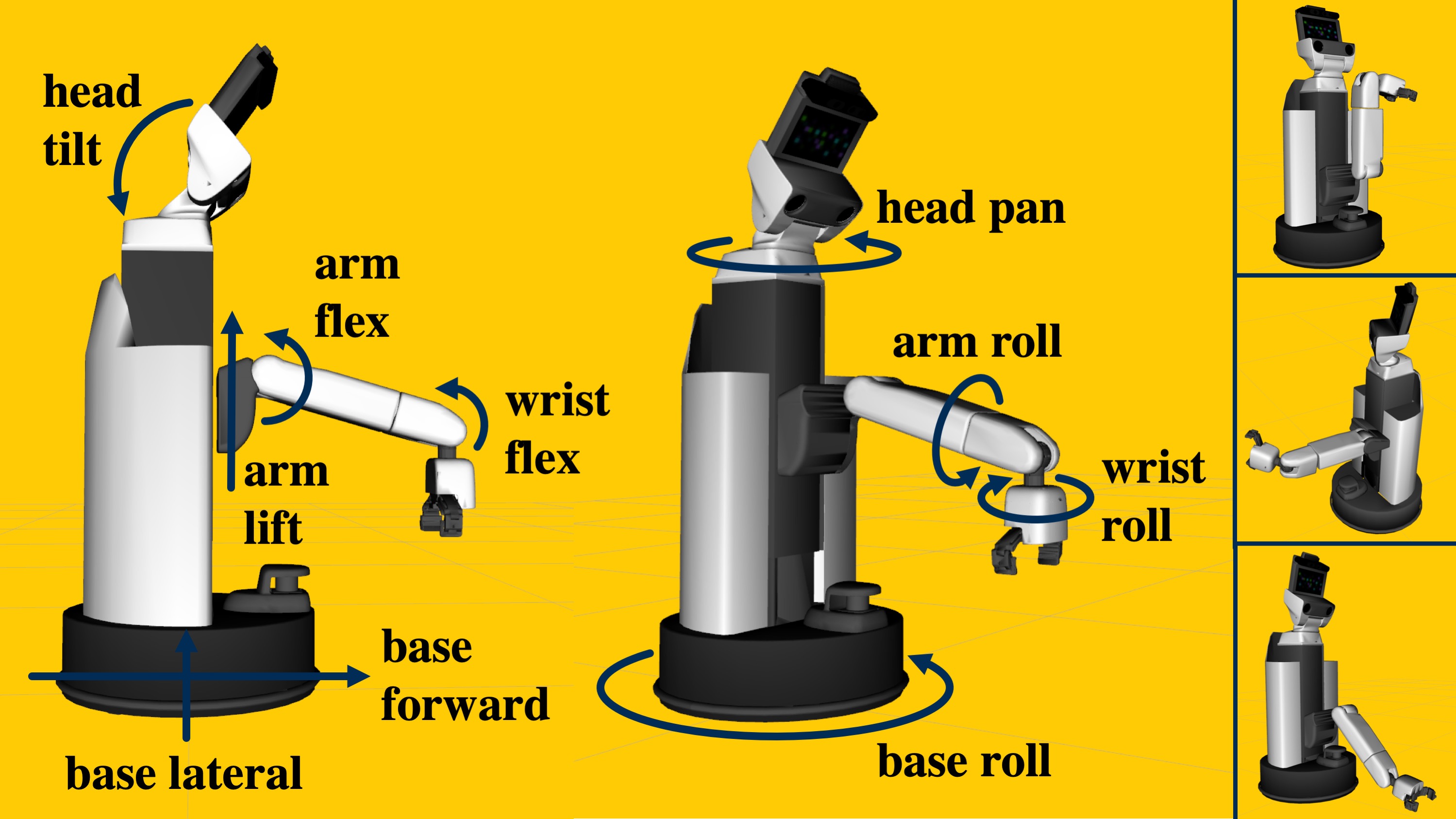}
	\caption{\textbf{HSR Control Model.} 
	}
	\label{fig:model}
\end{figure}

For our robot experiments, we use a Toyota Human Support Robot (HSR), which has a 4-DOF manipulator arm mounted on a torso with prismatic and revolute joints and a differential drive base \cite{UiYamaguchi2015,HSR_journal}.
Using the revolute joint atop its differential drive base, we effectively control HSR as an omnidirectional robot.
For visual servo control, we use the actuators shown in Figure~\ref{fig:model} as the joint space $\mathbf{q} \in \mathbb{R}^{10}$,
\begin{align}
\mathbf{q} = \begin{bmatrix} 
q_{\text{head tilt}}, q_{\text{head pan}}, \cdots, q_{\text{base roll}}
\end{bmatrix}^\intercal.
\label{eq:q}
\end{align}
In addition to $\mathbf{q}$, HSR's end effector has a parallel gripper with series elastic fingertips for grasping objects; the fingertips have 135~\textrm{mm} maximum width.

For perception, we use HSR's base-mounted UST-20LX 2D scanning laser for obstacle avoidance and the head-mounted Xtion PRO LIVE RGBD camera and end effector-mounted wide-angle grasp camera for segmentation.
The head tilt and pan joints act as a 2-DOF gimbal for the head camera, and the grasp camera moves with the arm and wrist joints; both cameras stream 640$\times$480 RGB images.

A significant component of HSR's manipulation DOF comes from its mobile base.
While many planning algorithms work well on high DOF arms with a stationary base, the odometer errors of HSR compound during trajectory execution and cause missed grasps.
Thus, VS is well-suited for HSR and other mobile robots, providing visual feedback on an object's relative position during mobile manipulation.


\section{Segmentation-based Visual Servo Control}


\subsection{Segmentation-based Features}
\label{sec:feat}

Assume we are given an RGB image $I$ containing an object of interest. Using VOS, we generate a binary mask
\begin{align}
M = \text{vos}(I,\mathbf{W}),
\label{eq:m}
\end{align}
where $M$ consists of pixel-level labels $\ell_p \in \{0,1\}$, $\ell_p=1$ indicates pixel $p$ corresponds to the segmented object, and $\mathbf{W}$ are learned VOS parameters (details in Section~\ref{sec:resseg}).

Using $M$, we define the following VOS-based features
\begin{align}
\label{eq:sa}
s_A := & ~\sum_{\ell_p \in M} \ell_p \\
s_x :=  & ~\frac{\sum_{\ell_p \in M, ~\ell_p = 1} p_x}{s_A} \label{eq:sx} \\
s_y := & ~\frac{\sum_{\ell_p \in M, ~\ell_p = 1} p_y}{s_A},
\label{eq:sy}
\end{align}
where $s_A$ is a measure of segmentation area by the number of labeled pixels, $s_x$ is the $x$-centroid of the segmented object using $x$-axis label positions $p_x$, and $s_y$ is the equivalent $y$-centroid.
In addition to \eqref{eq:sa}-\eqref{eq:sy}, we introduce more VOS features for depth estimation and grasping in Sections~\ref{sec:ade}-\ref{sec:grasp}.

\subsection{Visual Servo Control}
\label{sec:vs}

Using VOS-based features for our visual servo control scheme,  we define image feature error
\begin{align}
\mathbf{e}  := \mathbf{s} (I,\mathbf{W}) - \mathbf{s}^*,
\label{eq:e}
\end{align}
where $\mathbf{s} \in \mathbb{R}^k$ is the vector of visual features found in image $I$ using learned VOS parameters $\mathbf{W}$ and $\mathbf{s}^* \in \mathbb{R}^k$ is the vector of desired feature values.
In contrast to many VS control schemes, $\mathbf{e}$ in \eqref{eq:e} has no dependence on time, previous observations, or additional system parameters (e.g., camera parameters or 3D object models). 


Typical VS approaches relate camera motion to $\mathbf{s}$ using
\begin{align}
\label{eq:sdot}
\dot{\mathbf{s}} = \mathbf{L_s} \mathbf{v_c},
\end{align}
where $\mathbf{L_s} \in \mathbb{R}^{k \times 6}$ is a feature Jacobian relating the three linear and three angular camera velocities $\mathbf{v_c} \in \mathbb{R}^{6}$ to $\dot{\mathbf{s}}$.
From \eqref{eq:e}-\eqref{eq:sdot}, assuming ${\mathbf{\dot{s}}^*}=0 \implies \dot{\mathbf{e}}=\dot{\mathbf{s}} = \mathbf{L_s} \mathbf{v_c}$, we find the VS control velocities $\mathbf{v_c}$ to minimize $\mathbf{e}$ as
\begin{align}
\mathbf{v_c} = & ~\text{-} \lambda \widehat{\mathbf{L}_\mathbf{s}^+} \mathbf{e},
\label{eq:vc}
\end{align}
where $\widehat{\mathbf{L}_\mathbf{s}^+}$ is the estimated pseudoinverse of $\mathbf{L_s}$ and $\lambda$ ensures an exponential decoupled decrease of $\mathbf{e}$ \cite{ChHu06}.
Notably, VS control using \eqref{eq:vc} requires continuous, six degree of freedom (DOF) control of camera velocity.

To make \eqref{eq:vc} more general for discrete motion planning and fewer required control inputs, we modify \eqref{eq:sdot}-\eqref{eq:vc} to  
\begin{align}
\Delta \mathbf{s} =  &~\mathbf{J}_\mathbf{s} \Delta \mathbf{q} \\
\Delta \mathbf{q} = &~\text{-} \lambda \widehat{\mathbf{J}_\mathbf{s}^+} \mathbf{e},
\label{eq:Dq}
\end{align}
where $\Delta \mathbf{q}$ is the change of $\mathbf{q} \in \mathbb{R}^{n}$ actuated joints,
$\mathbf{J_s}\in\mathbb{R}^{k \times n}$ is the feature Jacobian relating $\Delta \mathbf{q}$  to $\Delta \mathbf{s}$, and $\widehat{\mathbf{J}_\mathbf{s}^+}$ is the estimated pseudoinverse of $\mathbf{J_s}$.
We command $\Delta \mathbf{q}$ directly to the robot joint space as our VOS-VS controller to minimize $\mathbf{e}$ and reach the desired feature values $\mathbf{s}^*$ in \eqref{eq:e}.

\subsection{Hadamard-Broyden Update Formulation}
\label{sec:hb}

In real visual servo systems, it is impossible to know the exact feature Jacobian ($\mathbf{J_s}$) relating control actuators to image features \cite{ChHu06}. Instead, some VS methods estimate $\mathbf{J_s}$ directly from observations \cite{ChHu07}; among these, a few use the Broyden update rule \cite{HoAs94, JaFuNe97,PiMcLi04}, which iteratively updates online.
In contrast to previous VS work, Broyden's original paper provides a formulation to estimate the \textit{pseudoinverse} feature Jacobian ($\widehat{\mathbf{J}_\mathbf{s}^+}$)  \cite[(4.5)]{Br65}.
However, we found it necessary to augment Broyden's formulation with a logical matrix $\mathbf{H}$, and define our new Hadamard-Broyden update
\begin{align}
\widehat{\mathbf{J}_\mathbf{s}^+}_{t+1} := \widehat{\mathbf{J}_\mathbf{s}^+}_t + \alpha \Bigg( \frac{ \big(\Delta \mathbf{q} - \widehat{\mathbf{J}_\mathbf{s}^+}_t \Delta \mathbf{e} \big) \Delta \mathbf{q}^\intercal \widehat{\mathbf{J}_\mathbf{s}^+}_t}{ \Delta \mathbf{q}^\intercal \widehat{\mathbf{J}_\mathbf{s}^+}_t \Delta \mathbf{e} } \Bigg) \circ \mathbf{H},
\label{eq:hb}
\end{align}
where $\alpha$ determines the update speed, 
$\Delta \mathbf{q}=\mathbf{q}_t - \mathbf{q}_{t-1}$ and $\Delta \mathbf{e}=\mathbf{e}_t - \mathbf{e}_{t-1}$ are the changes in joint space and feature errors since the last update, 
and $\mathbf{H} \in \mathbb{R}^{n \times k}$ is a logical matrix coupling actuators to image features.
In experiments, we initialize \eqref{eq:hb} using $\alpha=0.1$ and $\widehat{\mathbf{J}_\mathbf{s}^+}_{t=0}=0.001 \mathbf{H}$.

The Hadamard product with $\mathbf{H}$ prevents undesired coupling between certain actuator and image feature pairs.
In practice, we find that using the original Broyden update results in unpredictable convergence and learning gains for actuator-image feature pairs that are, in fact, unrelated.
Fortunately, we find that using $\mathbf{H}$ in \eqref{eq:hb} enables real-time convergence without any calibration on the robot for all of the experiment configurations in Section~\ref{sec:resvs}.




\subsection{VOS-VS Configurations}

We learn seven unique VOS-VS configurations using our HB update.
Using $s_x$ \eqref{eq:sx} and $s_y$ \eqref{eq:sy} in $\mathbf{e}$ \eqref{eq:e}, we define error 
\begin{align}
\mathbf{e}_{x,y}  := \mathbf{s}_{x,y} (M(I,\mathbf{W})) - \mathbf{s}^* 
= \begin{bmatrix} s_x \\ s_y \end{bmatrix} - \mathbf{s}^*.
\label{eq:exy}
\end{align}
Using $\mathbf{e}_{x,y}$ and HSR joints $\mathbf{q}$ \eqref{eq:q}, we choose $\widehat{\mathbf{J}_{\mathbf{s}}^+}$ in \eqref{eq:hb} as
\begin{align}
\label{eq:js}
\widehat{\mathbf{J}_{\mathbf{s}}^+}  \approx
\frac{\partial \mathbf{q}}{\partial \mathbf{e}_{x,y}} =
\frac{\partial \mathbf{q}}{\partial \mathbf{s}_{x,y} } = 
\begin{bmatrix}
\frac{\partial q_{\text{head tilt}} }{\partial s_x } & \frac{\partial q_{\text{head tilt}} }{\partial s_y } \\
\frac{\partial q_{\text{head pan}} }{\partial s_x } & \frac{\partial q_{\text{head pan}} }{\partial s_y }\\
\vdots & \vdots \\
\frac{\partial q_{\text{base roll}} }{\partial s_x } & \frac{\partial q_{\text{base roll}} }{\partial s_y }
\end{bmatrix},
\end{align}
where $\widehat{\mathbf{J}_{\mathbf{s}}^+} \in \mathbb{R}^{10 \times 2}$.  
Note that in our Hadamard-Broyden update \eqref{eq:hb}, each element $\frac{\partial q_i}{\partial s_j}$ in $\widehat{\mathbf{J}_{\mathbf{s}}^+}$ is multiplied by element $\mathbf{H}_{i,j}$ in the Hadamard product.
Thus, we configure the logical coupling matrix $\mathbf{H}$ by setting $\mathbf{H}_{i,j}= 1$ if coupling actuated joint $q_i$ with image feature $s_j$ is desired.
Using our update formulation \eqref{eq:hb}, we learn $\widehat{\mathbf{J}_{\mathbf{s}}^+}$ on HSR for the seven $\mathbf{H}$ configurations listed in Table~\ref{table:lresults} and provide experimental results for each configuration in Section~\ref{sec:resvs}.

\setlength{\tabcolsep}{4pt} 
\begin{table}
	\centering
	\caption{
		\textbf{VOS-VS Hadamard-Broyden Update Configurations.} $\widehat{\mathbf{J}_{\mathbf{s}}^+}$ values are learned online using our HB update formulation \eqref{eq:hb}, enabling HSR to automatically learn the relationship between actuators and visual features without any camera calibration.}
	\footnotesize
	\begin{tabular}{ l |l | l | l | l | l}
		\hline 
		& & \multicolumn{4}{c}{} \\[-1em]
		\multicolumn{1}{c|}{$\mathbf{H}$ \eqref{eq:hb}} & & \multicolumn{4}{c}{Learned $\frac{\partial q_i}{\partial s_j}$ in $\widehat{\mathbf{J}_{\mathbf{s}}^+}$ \eqref{eq:js}} \\
		\cline{3-6}
		\multicolumn{1}{c|}{Config.} & Camera & \multicolumn{2}{c|}{$s_x$} & \multicolumn{2}{c}{$s_y$} \\
		\hline
		$\mathbf{H}_{\text{head}}$	&	Head	&	$q_{\text{head pan}}$ & ~0.00173	&	$q_{\text{head tilt}}$ & 0.00183		\\
		\rowcolor{rowgray}	$\mathbf{H}_{\text{arm lift}}$	&	Grasp	&	$q_{\text{arm lift}}$ & -0.00157	&	$q_{\text{arm roll}} $ &  0.00321		\\
		$\mathbf{H}_{\text{arm wrist}}$	&	Grasp	&	$q_{\text{wrist flex}} $ & -0.00221	&	$q_{\text{arm roll}} $ & 0.00445		\\
		\rowcolor{rowgray}		&		&	$q_{\text{arm lift}}$ & -0.00036	&	&		\\
		\rowcolor{rowgray}	$\mathbf{H}_{\text{arm both}}$	&	Grasp	&	 $q_{\text{wrist flex}}$ & -0.00392	&	$q_{\text{arm roll}} $ &  0.00328		\\
		$\mathbf{H}_{\text{base}}$	&	Grasp	&	$q_{\text{base forward}} $ &  -0.00179	&	$q_{\text{base lateral}} $ & 0.00173		\\
		\rowcolor{rowgray}	$\mathbf{H}_{\text{base grasp}}$	&	Grasp	&	$q_{\text{base forward}} $ &  -0.00040	&	$q_{\text{base lateral}} $ &  0.00040		\\
		\hline
	\end{tabular}
	\label{table:lresults}
\end{table}
\begin{figure}
	\centering
	\includegraphics[width=0.4625\textwidth]{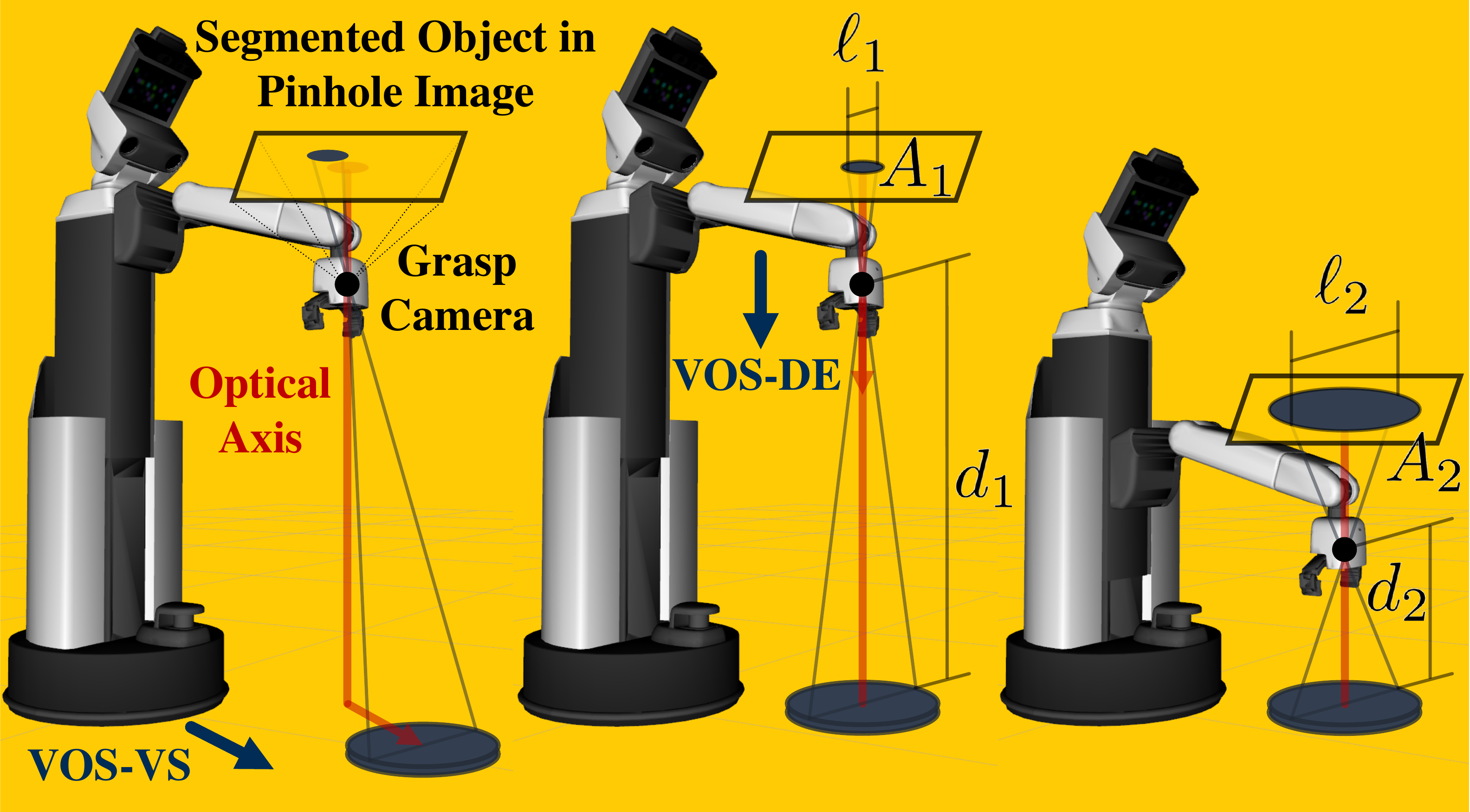}
	\caption{
		\textbf{VOS-based Visual Servo and Depth Estimation.}
		HSR first aligns an object with the camera's optical axis then estimates the object's depth as the camera approaches.
		Using Galileo's Square-cube law \eqref{eq:sc}, we estimate the object's depth using changes in relative camera position and segmentation area.
	}
	\label{fig:vsade}
\end{figure}

\section{Segmentation-based Depth Estimation}
\label{sec:ade}

\begin{figure*}
	\centering
	\includegraphics[width=0.975\textwidth]{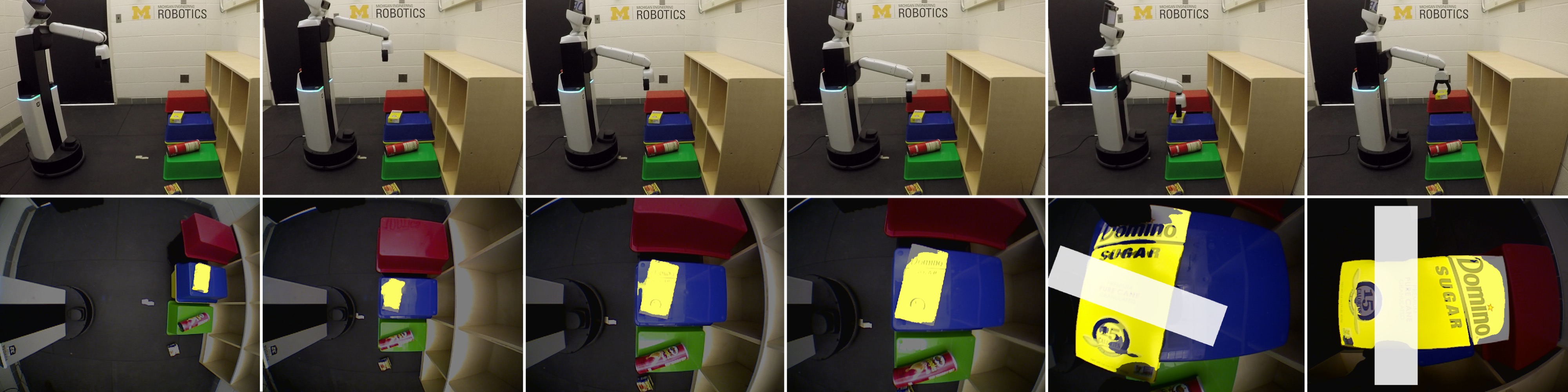}
	\caption{\textbf{VOS-based Grasping}. VOS-based visual servo control (columns 1 to 2), active depth estimation (2-4), and mobile robot grasping (5-6). Using our combined framework with a single RGB camera, HSR identifies the sugar box, locates it in 3D, and picks it up in real-time.
}
	\label{fig:complete}
\end{figure*}

By combining VOS-based features with active perception, we are able to estimate the depth of segmented objects 
and approximate their 3D position.
As shown in Figure~\ref{fig:vsade}, we initiate our depth estimation framework (VOS-DE) by centering the optical axis of our camera with a segmented object using the $\mathbf{H}_{\text{base}}$ VOS-VS controller.
This alignment minimizes lens distortion, which facilitates the use of an ideal camera model.
Using the pinhole camera model \cite{FoPo02}, projections of objects onto the image plane scale inversely with their distance on the optical axis from the camera. 

Thus, with the object centered on the optical axis, we can relate projection scale and object distance using
\begin{align}
\ell_1 d_1 = \ell_2 d_2 \implies 
\frac{\ell_2}{\ell_1} = \frac{d_1}{d_2},
\label{eq:ph}
\end{align}
where $\ell_1$ is the projected length of an object measurement orthogonal to the optical axis, $d_1$ is the distance along the optical axis of the object away from the camera, and $\ell_2$ is the projected measurement length at a new distance $d_2$. 
Combining Galileo Galilei's Square-cube law with \eqref{eq:ph}, 
\begin{align}
A_2 = A_1 \bigg(\frac{\ell_2}{\ell_1}\bigg)^2 \implies
A_2  = A_1 \bigg(\frac{d_1}{d_2}\bigg)^2,
\label{eq:sc}
\end{align}
where $A_1$ is the projected object area corresponding to $\ell_1$ and $d_1$ (see Figure~\ref{fig:vsade}).
As the camera advances on the optical axis, we modify \eqref{eq:sc} to relate collected images using
\begin{align}
d_1 \sqrt{A_1} = d_2 \sqrt{A_2} = c_{\text{object}},
\label{eq:cs}
\end{align}
where $c_{\text{object}}$ is a constant proportional to the orthogonal surface area of the segmented object.
Also, using a coordinate frame with the $z$ axis aligned with the optical axis, 
\begin{align}
d = z_{\text{camera}} - z_{\text{object}},
\label{eq:d}
\end{align}
where $z_{\text{camera}}$ and $z_{\text{object}}$ are the $z$-axis coordinates of the camera and object.
Because the camera and object are both centered on the $z$ axis, $x_{\text{camera}}=x_{\text{object}}=0$ and $y_{\text{camera}}=y_{\text{object}}=0$.
Using \eqref{eq:d} and $s_A$ \eqref{eq:sa}, we update \eqref{eq:cs} as 
\begin{align}
\nonumber
(z_{\text{camera},1} - z_{\text{object}}) \sqrt{s}_{A,1} 
= &~(z_{\text{camera},2} - z_{\text{object}}) \sqrt{s}_{A,2} \\ = &~c_{\text{object}},
\label{eq:measure}
\end{align}
where the object is assumed stationary between images (i.e., $\dot{z}_{\text{object}}=0$) and the $z_{\text{camera}}$ position is known from the robot's kinematics. 
Note that $z_{\text{camera}}$ provides relative depth for VOS-DE and \eqref{eq:measure} identifies a key linear relationship between $\sqrt{s}_{A}$ and the distance between the object and camera.

Finally, after collecting a series of $m$ measurements, we estimate the depth of the segmented object.
From \eqref{eq:measure},
\begin{align}
z_{\text{object}} \sqrt{s}_{A,1} + c_{\text{object}} = z_{\text{camera},1} \sqrt{s}_{A,1},
\end{align}
which over the $m$ measurements in $\mathbf{A}\mathbf{x}=\mathbf{b}$ form yields
\begin{align}
\begin{bmatrix}
\sqrt{s}_{A,1} & 1 \\ \sqrt{s}_{A,2} & 1 \\ \vdots & \vdots \\ \sqrt{s}_{A,m} & 1
\end{bmatrix}
\begin{bmatrix}
\hat{z}_{\text{object}} \\ \hat{c}_{\text{object}}\\ 
\end{bmatrix}
=
\begin{bmatrix}
z_{\text{camera},1} \sqrt{s}_{A,1} \\ z_{\text{camera},2} \sqrt{s}_{A,2} \\ \vdots \\ z_{\text{camera},m} \sqrt{s}_{A,m}
\end{bmatrix}.
\label{eq:axb}
\end{align}
By solving \eqref{eq:axb} for $\hat{z}_{\text{object}}$ and  $\hat{c}_{\text{object}}$, we estimate the distance $d$ in \eqref{eq:d}, and, thus, the 3D location of the object.
In Section~\ref{sec:vsde}, we show that our combined VOS-VS and VOS-DE framework is sufficient for locating, approaching, and estimating the depth of a variety of unstructured objects.

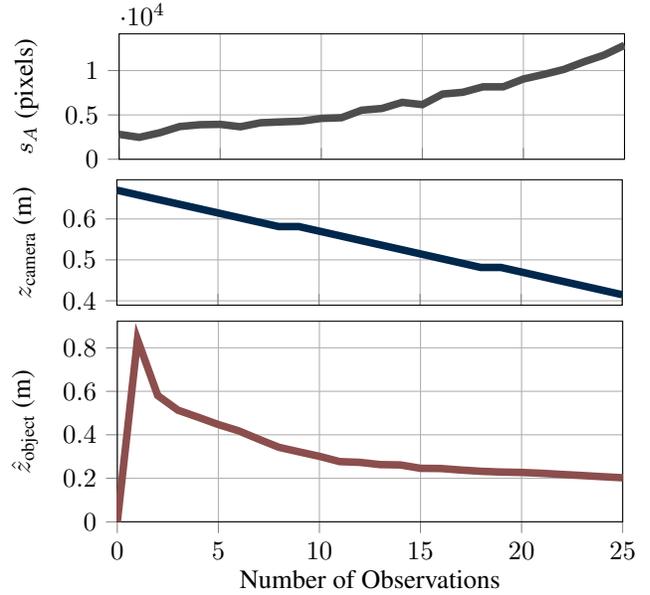
\begin{figure}
	\noindent\begin{minipage}{0.475\textwidth}
		\centering
\begin{tikzpicture}

\definecolor{color0}{rgb}{0.12156862745098,0.466666666666667,0.705882352941177}
\definecolor{color0}{rgb}{0.00000,0.15290,0.29800} 
\definecolor{color0}{rgb}{0.3,0.3,0.3}


\begin{axis}[
width=8.3cm,
height=3.25cm,
tick align=outside,
tick pos=left,
x grid style={lightgray!92.026143790849673!black},
xmajorgrids,
xmin=0, xmax=37,
xmin=0, xmax=25,
y grid style={lightgray!92.026143790849673!black},
ylabel={$s_A$ (pixels)},
ymajorgrids,
xtick align = inside,
xticklabels={},
grid style={lightgray!50},
ymin=0,
/tikz/inner sep to outer sep/.style={inner sep=0pt, outer sep=.3333em},
x tick label style=inner sep to outer sep,
x label style=inner sep to outer sep,
y label style=inner sep to outer sep,
]
\addplot [line width = 1mm, color0, forget plot]
table [row sep=\\]{%
0	2810 \\
1	2472 \\
2	2981 \\
3	3688 \\
4	3895 \\
5	3935 \\
6	3663 \\
7	4116 \\
8	4210 \\
9	4292 \\
10	4604 \\
11	4683 \\
12	5531 \\
13	5726 \\
14	6413 \\
15	6159 \\
16	7351 \\
17	7552 \\
18	8166 \\
19	8168 \\
20	9049 \\
21	9561 \\
22	10127 \\
23	10989 \\
24	11759 \\
25	12852 \\
};

\end{axis}

\end{tikzpicture}
	\end{minipage}%
	\hfill
	\noindent\begin{minipage}{0.472\textwidth}
		\centering
\begin{tikzpicture}

\definecolor{color0}{rgb}{0.12156862745098,0.466666666666667,0.705882352941177}
\definecolor{color0}{rgb}{0.00000,0.15290,0.29800} 

%
%

\begin{axis}[
width=8.3cm,
height=3.25cm,
tick align=outside,
tick pos=left,
x grid style={lightgray!92.026143790849673!black},
xmajorgrids,
xmin=0, xmax=37,
xmin=0, xmax=25,
y grid style={lightgray!92.026143790849673!black},
ylabel={$z_{\text{camera}}$ (\textrm{m})},
ymajorgrids,
xticklabels={},
xtick align = inside,
grid style={lightgray!50},
/tikz/inner sep to outer sep/.style={inner sep=0pt, outer sep=.3333em},
x tick label style=inner sep to outer sep,
x label style=inner sep to outer sep,
y label style=inner sep to outer sep,
]
\addplot [line width = 1mm, color0, forget plot]
table [row sep=\\]{%
0	0.670346668663 \\
1	0.659235557552 \\
2	0.64812444644 \\
3	0.637013335329 \\
4	0.625902224218 \\
5	0.614791113107 \\
6	0.603680001996 \\
7	0.592568890885 \\
8	0.581457779774 \\
9	0.581464936433 \\
10	0.570353825322 \\
11	0.559242714211 \\
12	0.548131603099 \\
13	0.537020491988 \\
14	0.525909380877 \\
15	0.514798269766 \\
16	0.503687158655 \\
17	0.492576047544 \\
18	0.481464936433 \\
19	0.481477307392 \\
20	0.470366196281 \\
21	0.45925508517 \\
22	0.448143974059 \\
23	0.437032862947 \\
24	0.425921751836 \\
25	0.414810640725 \\
};

\end{axis}

\end{tikzpicture}
	\end{minipage}%
	\hfill
	\noindent\begin{minipage}{0.475\textwidth}
		\centering
\begin{tikzpicture}

\definecolor{color0}{rgb}{0.12156862745098,0.466666666666667,0.705882352941177}


\definecolor{color0}{rgb}{0.00000,0.15290,0.29800} 
\definecolor{color0}{rgb}{0.25,0.25,0.25} 
\definecolor{color0}{rgb}{0.55,0.3,0.3}

\begin{axis}[
width=8.3cm,
height=3cm,
height=4.25cm,
tick align=outside,
tick pos=left,
x grid style={lightgray!92.026143790849673!black},
xlabel={Number of Data Points ($m$)},
xlabel={Number of Observations},
ylabel={$\hat{z}_{\text{object}}$ (\textrm{m})},
xmajorgrids,
xmin=0, xmax=37,
xmin=0, xmax=25,
y grid style={lightgray!92.026143790849673!black},
ymajorgrids,
grid style={lightgray!50},
ymin=0,
/tikz/inner sep to outer sep/.style={inner sep=0pt, outer sep=.3333em},
x tick label style=inner sep to outer sep,
x label style=inner sep to outer sep,
y label style=inner sep to outer sep,
]
\addplot [line width = 1mm, color0, forget plot]
table [row sep=\\]{%
0	0 \\
1	0.838248944173827 \\
2	0.580596324824258 \\
3	0.51327782252781 \\
4	0.480736177375978 \\
5	0.447146039076824 \\
6	0.418321847008474 \\
7	0.3795574641873 \\
8	0.342407326003592 \\
9	0.32146351509547 \\
10	0.301008729583142 \\
11	0.276785857363247 \\
12	0.273291394348314 \\
13	0.263023004339671 \\
14	0.261461772420796 \\
15	0.245963297275762 \\
16	0.245076458868052 \\
17	0.238018224144223 \\
18	0.232195709107898 \\
19	0.228676824835925 \\
20	0.226921659776938 \\
21	0.223014096115807 \\
22	0.217778844786591 \\
23	0.212836451281498 \\
24	0.207393845137173 \\
25	0.202500914277393 \\
};

\end{axis}

\end{tikzpicture}
	\end{minipage}%
	\caption{\textbf{
			Depth Estimate of Sugar Box.} Data collected and processed in real-time during the initial approach in Figure~\ref{fig:complete}.
	}
	\label{fig:de}
\end{figure}

\noindent \textbf{Remark}:  There are many methods to find approximate solutions to \eqref{eq:axb}. In practice, we find that a least squares solution provides robustness to outliers caused by segmentation errors (see visual and quantitative example in Figures~\ref{fig:complete}-\ref{fig:de}).

\section{Segmentation-based Grasping}
\label{sec:grasp}

We develop a VOS-based method of grasping and grasp-error detection (VOS-Grasp).
Assuming an object is centered and has estimated depth $\hat{z}_{\text{object}}$, we move $z_{\text{camera}}$ to
\begin{align}
z_{\text{camera, grasp}} = \hat{z}_{\text{object}} + z_{\text{gripper}},
\label{eq:grasp}
\end{align}
where $z_{\text{gripper}}$ is the known $z$-axis offset between $z_{\text{camera}}$ and the center of HSR's closed fingertips. Thus, when $z_{\text{camera}}$ is at $z_{\text{camera, grasp}}$, HSR can reach the object at depth $\hat{z}_{\text{object}}$.

After moving to $z_{\text{camera, grasp}}$, we center the object directly underneath HSR's antipodal gripper using $\mathbf{H}_{\text{base grasp}}$ VOS-VS control.
To find a suitable grasp location, we project and rotate a mask of the gripper, $M_{\text{grasp}}$, into the camera as shown in column 5 of Figure~\ref{fig:complete} and solve
\begin{align}
\arg \min_{q_{\text{wrist roll}}}  \mathcal{J}(q_{\text{wrist roll}}) =\frac{M\cap M_{\text{grasp}}(q_{\text{wrist roll}})}{M \cup M_{\text{grasp}}(q_{\text{wrist roll}})},
\label{eq:jwrist}
\end{align}
where $\mathcal{J}$ is the intersection over union (or Jaccard index \cite{jaccard}) of $M_{\text{grasp}}$ and object segmentation mask $M$, and $M_{\text{grasp}}(q_{\text{wrist roll}})$ is the projection of $M_{\text{grasp}}$ corresponding to HSR wrist rotation $q_{\text{wrist roll}}$.
Thus, we grasp the object using the wrist rotation with least intersection between the object and the gripper, which is then less likely to collide with the object before achieving a parallel grasp. 

After the object is grasped, we lift HSR's arm to perform a visual grasp check. 
We consider a grasp complete if
\begin{align}
s_{A, \text{raised}} > 0.5 ~ s_{A, \text{grasp}},
\end{align}
where $s_{A, \text{grasp}}$ is the object segmentation size $s_A$ \eqref{eq:sa} during the initial grasp and $s_{A, \text{raised}}$ is the corresponding $s_A$ after lifting the arm.
If $s_A$ decreases when lifting the arm, the object is further from the camera and not securely grasped.
Thus, we quickly identify if a grasp is missed and regrasp as necessary.
Note that this VOS-based grasp check can also work with other grasping methods \cite{GuPaSaPl2016,mahler2017dex}.
A complete demonstration of our VOS-based visual servo control, depth estimation, and grasping framework is shown in Figure~\ref{fig:complete}. 

\begin{figure}
	\centering
	\includegraphics[width=0.4625\textwidth]{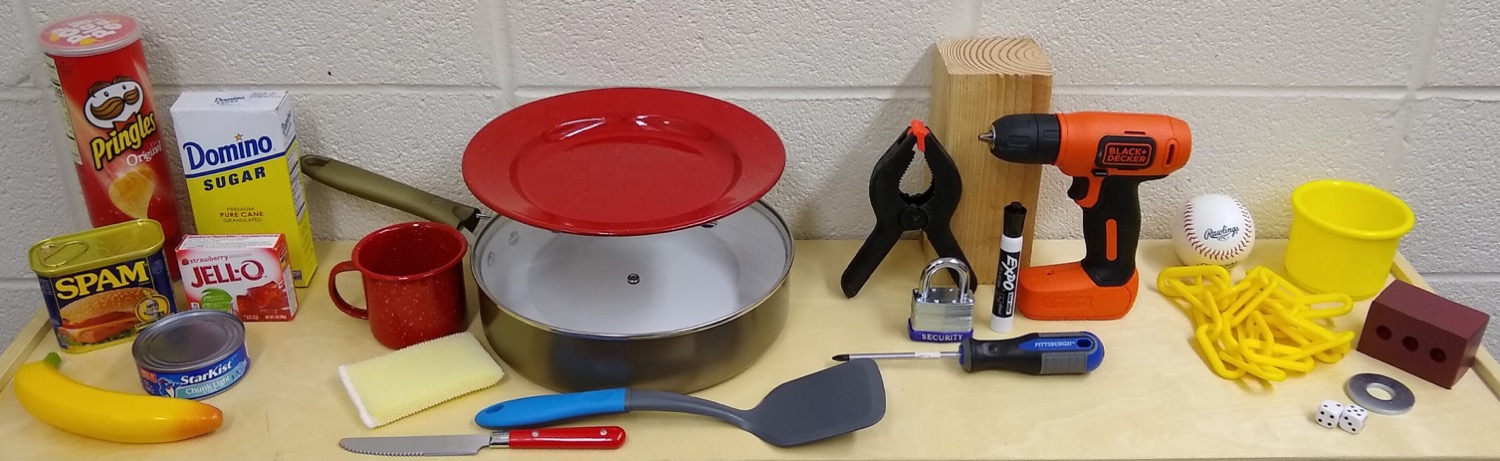}
	\caption{\textbf{Experiment Objects from YCB Dataset.} Object categories are (from left to right) Food, Kitchen, Tool, and Shape. Spanning from 470~\textrm{mm} long to the 4~\textrm{mm} thick, we intentionally select many of the challenge objects to break our framework.} 
	\label{fig:objects}
\end{figure}

\section{ROBOT EXPERIMENTS}
\label{sec:exp}

\subsection{Experiment Objects}
\label{sec:resobj}

For most of our experiments, we use the objects from the YCB object dataset \cite{YCB} shown in Figure~\ref{fig:objects}.
We use six objects from each of the food, kitchen, tool, and shape categories and purposefully choose some of the most difficult objects.
To name only a few of the challenges for the selected objects: dimensions span from the 470~\textrm{mm} long pan to the 4~\textrm{mm} thick washer, most of the contours change with pose, and over a third of the objects exhibit specular reflection of overhead lights.
To learn object recognition, we annotate ten training images of each object using HSR's grasp camera with various object poses, backgrounds, and distances from the camera (see example image in Figure~\ref{fig:stairs}).

\begin{figure}
	\centering
	\includegraphics[width=0.235\textwidth]{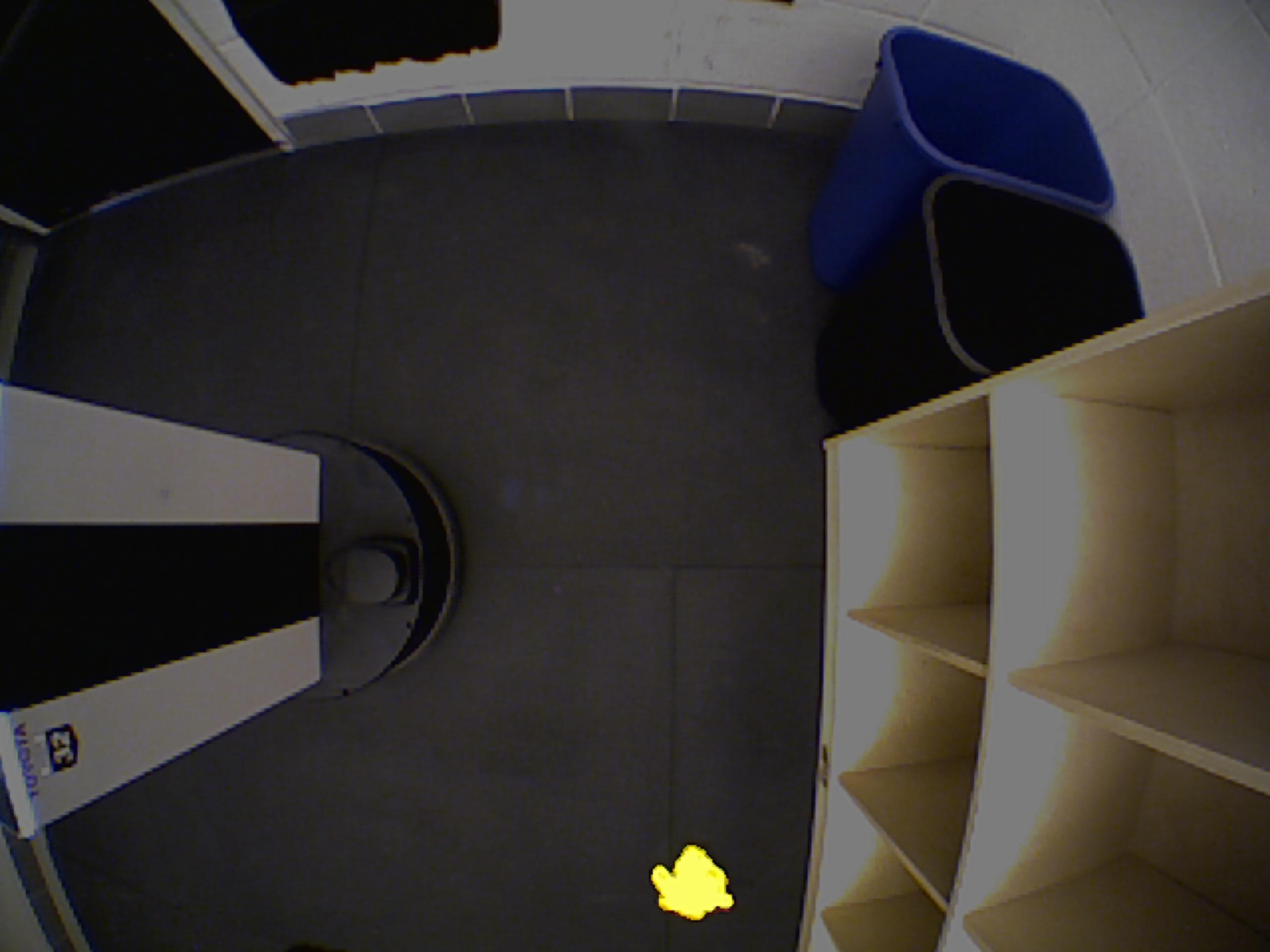}
\begin{tikzpicture}

\definecolor{color1}{rgb}{0.3,1,0.3}
\definecolor{colorbanana}{rgb}{1,0.9,0}
\definecolor{colorwood}{rgb}{0.86,0.71,0.52}
\definecolor{colorwasher}{rgb}{0.86,0.89,0.92}
\definecolor{colorwasher}{rgb}{0.72,0.78,0.84}
\definecolor{colorplate}{rgb}{0.6,0,0}

\definecolor{color0}{rgb}{0.705882352941177,0.466666666666667,0.12156862745098}
\definecolor{colorblue}{rgb}{0.00000,0.15290,0.29800} 
\definecolor{colorblue}{rgb}{0.00000,0.15290,0.29800} 
\definecolor{colorhb}{rgb}{0.3,0.85,0.3} 
\definecolor{colorred}{rgb}{1,0.1,0.1}

\begin{axis}[
ticks = none,
tick align=outside,
tick pos=left,
width=5.7cm,
height=4.6675cm,
x grid style={lightgray!92.026143790849673!black},
xmajorgrids,
xmin=0, xmax=640,
xtick={0,160,320,480,640},
y grid style={lightgray!92.026143790849673!black},
grid style={lightgray!50},
legend style={at={(0.015,0.98)},anchor=north west},
legend cell align=left,
xticklabels={},
yticklabels={},
ymin=-480, ymax=0,
ytick={0,-120,-240,-360,-480},
ymajorgrids,
/tikz/inner sep to outer sep/.style={inner sep=0pt, outer sep=.3333em},
x tick label style=inner sep to outer sep,
x label style=inner sep to outer sep,
y label style=inner sep to outer sep,
]

\addplot [line width = 0.75mm, colorred]
table [row sep=\\]{%
	346.55624757	-446.33974609 \\
	355.26494362	-402.786727365 \\
	364.976557355	-371.555987916 \\
	380.924488361	-334.390482751 \\
	405.507062423	-298.712062161 \\
	444.902071108	-255.62529175 \\
	508.579767734	-224.058206829 \\
	574.326047359	-225.163934426 \\
};
\addlegendentry{\footnotesize Original Update};

\addplot [line width = 0.75mm, colorblue]
table [row sep=\\]{%
	349.663396122	-445.087877312 \\
	358.395127958	-400.286326276 \\
	367.688846968	-372.31581741 \\
	381.453249282	-338.665854568 \\
	398.209139198	-299.268879283 \\
	421.270334727	-271.139477735 \\
	459.803933632	-254.050005042 \\
	441.816916106	-248.340913477 \\
	436.282659101	-241.846473908 \\
	432.629489912	-240.655858632 \\
	422.869313437	-236.116703035 \\
	388.77318485	-234.833305573 \\
	346.135614582	-237.77617131 \\
	329.844787641	-238.469818759 \\
	324.598409799	-238.665608601 \\
	322.286473271	-238.610694138 \\
	320.371085694	-238.30524734 \\
};
\addlegendentry{\footnotesize Ours \eqref{eq:hb}};

\node[right, align=left, inner sep=0mm, text=black]
at (axis cs:175,-460,0) {\scriptsize Initial Target Location};

\node[right, align=left, inner sep=0mm, text=black]
at (axis cs:175,-200,0) {\scriptsize Centered on Target};

\node[right, align=left, inner sep=0mm, text=black]
at (axis cs:515,-262,0) {\scriptsize (Crash)};

\addplot [only marks,line width=0.75pt,mark size=3.5pt,mark=*,mark options={solid}, color1]
table [row sep=\\]{%
	320	-240 \\
};

\end{axis}
\end{tikzpicture}
	\caption{\textbf{Learning $\widehat{\mathbf{J}_\mathbf{s}^+}$ for $\mathbf{H}_{\text{base}}$}. Visual servo trajectory of the target object in image space (right) using the original Broyden update (red) and our Hadamard-Broyden update \eqref{eq:hb} (blue). Starting with the same $\widehat{\mathbf{J}_\mathbf{s}^+}_{t=0}$ and offset target location (yellow chain, left), the original update leads HSR into the wall while our update learns the correct visual servoing parameters to center HSR on the target.
	}
	\label{fig:vs_init}
\end{figure}
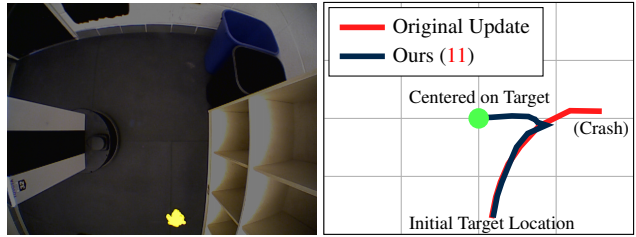

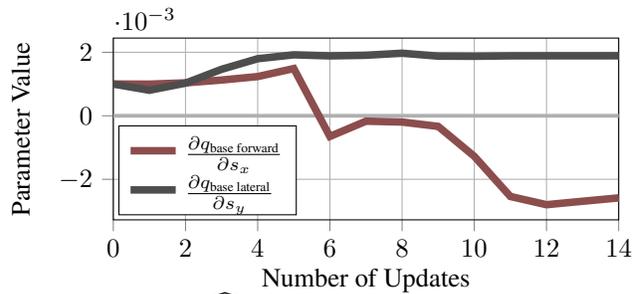
\begin{figure}
	\centering
\begin{tikzpicture}

\definecolor{color0}{rgb}{0.12156862745098,0.466666666666667,0.705882352941177}
\definecolor{color0}{rgb}{0.55,0.3,0.3}
\definecolor{color1}{rgb}{0.00000,0.15290,0.29800} 
\definecolor{color1}{rgb}{0.3,0.3,0.3} 


\begin{axis}[
width=8.3cm,
height=4cm,
tick align=outside,
tick pos=left,
x grid style={lightgray!92.026143790849673!black},
xlabel={Number of Updates},
xmajorgrids,
xmin=0, xmax=14,
y grid style={lightgray!92.026143790849673!black},
ylabel={Parameter Value},
ymajorgrids,
grid style={lightgray!50},
legend style={at={(0.01,0.025)},anchor=south west},
legend cell align=left,
/tikz/inner sep to outer sep/.style={inner sep=0pt, outer sep=.3333em},
x tick label style=inner sep to outer sep,
x label style=inner sep to outer sep,
y label style=inner sep to outer sep,
]

\addplot [line width = 0.5mm, lightgray!92.026143790849673!black, forget plot]
table [row sep=\\]{
0 0 \\
14 0 \\
};

\addplot [line width = 1mm, color0]
table [row sep=\\]{%
0	0.001 \\
1	0.0009959 \\
2	0.00103736 \\
3	0.00112583 \\
4	0.00123497 \\
5	0.00148743 \\
6	-0.00065174 \\
7	-0.00017049 \\
8	-0.00019697 \\
9	-0.00033457 \\
10	-0.00127627 \\
11	-0.00254168 \\
12	-0.00280058 \\
13	-0.00269248 \\
14	-0.00258629 \\
};
\addlegendentry{$\frac{\partial q_{\text{base forward}}}{\partial s_x}$};

\addplot [line width = 1mm, color1]
table [row sep=\\]{%
	0	0.001 \\
	1	0.0008041 \\
	2	0.00102736 \\
	3	0.00146392 \\
	4	0.00179695 \\
	5	0.00192028 \\
	6	0.00188602 \\
	7	0.001906 \\
	8	0.00197022 \\
	9	0.00187844 \\
	10	0.00187367 \\
	11	0.00188735 \\
	12	0.00188883 \\
	13	0.00188871 \\
	14	0.00188944 \\
};
\addlegendentry{$\frac{\partial q_{\text{base lateral}}}{\partial s_y}$};

\end{axis}

\end{tikzpicture}
	\caption{\textbf{Learning $\widehat{\mathbf{J}_\mathbf{s}^+}$ Parameters for $\mathbf{H}_{\text{base}}$}. This plot corresponds to the fourteen Hadamard-Broyden updates used to learn visual servoing parameters in Figure~\ref{fig:vs_init}. 
		$\frac{\partial q_{\text{base forward}}}{\partial s_x}$ initializes with the incorrect sign but still converges using our update formulation.}
	\label{fig:init_param}
\end{figure}

\subsection{Video Object Segmentation Method}
\label{sec:resseg}

We segment objects using OSVOS \cite{OSVOS}. 
OSVOS uses a base network trained on ImageNet \cite{ImageNet} to recognize image features, re-trains a parent network on DAVIS \cite{DAVIS} to learn general video object segmentation, and then fine tunes for each of our experiment objects (i.e., each object has unique learned parameters $\mathbf{W}$ in \eqref{eq:m}).
After learning $\mathbf{W}$, our VOS framework segments HSR's 640$\times$480 RGB images at 29.6~\textrm{Hz} using a single GPU (GTX 1080 Ti).

\subsection{VOS-VS Results}
\label{sec:resvs}

\noindent\textbf{Hadamard-Broyden Update}
We learn all of the VOS-VS configurations in Table~\ref{table:lresults} on HSR using the Hadamard-Broyden update formulation in \eqref{eq:hb}.
We initialize each configuration using $\widehat{\mathbf{J}_\mathbf{s}^+}_{t=0}=0.001~\mathbf{H}$, $\alpha=0.1$, and a target object in view to elicit a step response from the VOS-VS controller (see Figure~\ref{fig:vs_init}).
Each configuration starts at a specific pose (e.g., $\mathbf{H}_{\text{base}}$ uses the leftmost pose in Figures~\ref{fig:vsade}-\ref{fig:complete}), and configurations use $s^* = [320, 240]'$ in \eqref{eq:exy}, except for $\mathbf{H}_{\text{base grasp}}$, which uses $s^* = [220, 240]'$ to position grasps.


When initializing each configuration, after a few iterations of control inputs from \eqref{eq:Dq} and updates from \eqref{eq:hb}, the learned $\widehat{\mathbf{J}_\mathbf{s}^+}$ matrix generally shows convergence for any $\mathbf{H}_{i,j}$ component that is initialized with the correct sign (e.g., five updates for $\frac{\partial q_{\text{base lateral}}}{\partial s_y}$ in Figure~\ref{fig:init_param}).
Components initialized with an incorrect sign generally require more updates to change directions and jump through zero during one of the discrete updates (e.g., $\frac{\partial q_{\text{base forward}}}{\partial s_x}$ in Figure~\ref{fig:init_param}).
If an object goes out of view from an incorrectly signed component, we reset HSR's pose and restart the update from the most recent  $\widehat{\mathbf{J}_\mathbf{s}^+}_{t}$.
Once $s^*$ is reached, the object can be moved to elicit a few more step responses for fine tuning.
Table~\ref{table:lresults} shows the learned parameters for each configuration.
In the remaining experiments, we set $\alpha=0$ in \eqref{eq:hb} to reduce variability.

\begin{figure}
	\centering
	\includegraphics[width=0.235\textwidth]{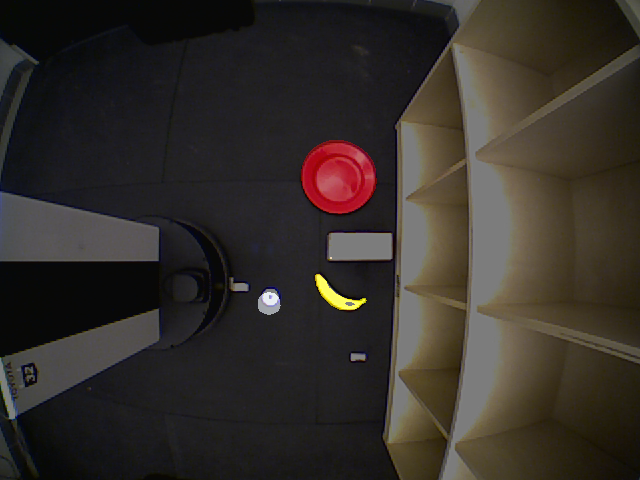}
\begin{tikzpicture}

\definecolor{color1}{rgb}{0.3,1,0.3}
\definecolor{colorbanana}{rgb}{1,0.9,0}
\definecolor{colorwood}{rgb}{0.86,0.71,0.52}
\definecolor{colorwasher}{rgb}{0.86,0.89,0.92}
\definecolor{colorwasher}{rgb}{0.72,0.78,0.84}
\definecolor{colorplate}{rgb}{0.6,0,0}

\begin{axis}[
ticks = none,
tick align=outside,
tick pos=left,
width=5.7cm,
height=4.6675cm,
x grid style={lightgray!92.026143790849673!black},
xmajorgrids,
xmin=0, xmax=640,
xtick={0,160,320,480,640},
y grid style={lightgray!92.026143790849673!black},
grid style={lightgray!50},
legend style={at={(0.02,0.02)},anchor=south west},
legend cell align=left,
xticklabels={},
yticklabels={},
ymin=-480, ymax=0,
ytick={0,-120,-240,-360,-480},
ymajorgrids,
/tikz/inner sep to outer sep/.style={inner sep=0pt, outer sep=.3333em},
x tick label style=inner sep to outer sep,
x label style=inner sep to outer sep,
y label style=inner sep to outer sep,
]
\addplot [only marks,line width=0.75pt,mark size=1.5pt,mark=*,mark options={solid}, colorwood]
table [row sep=\\]{%
331.834848485	-246.107575758 \\
};

\addplot [line width = 0.75mm, colorwasher, forget plot]
table [row sep=\\]{%
322.31791528	-214.868325445 \\
268.47220727	-304.292844686 \\
292.661470315	-275.527438626 \\
304.892234594	-260.82326028 \\
311.854959237	-251.690128052 \\
314.493223522	-248.058737542 \\
316.037928215	-245.773522452 \\
316.665236937	-244.751869682 \\
316.823632007	-244.281512393 \\
316.835277442	-244.224858136 \\
};

\addplot [line width = 0.75mm, colorplate, forget plot]
table [row sep=\\]{%
337.789988176	-177.049316117 \\
330.821857218	-204.777635777 \\
325.874268933	-220.27568695 \\
323.875191013	-228.949911592 \\
323.146844086	-233.920284146 \\
322.985604888	-236.274425686 \\
322.905551638	-237.064629583 \\
322.884118944	-237.048169515 \\
322.887695854	-237.039768072 \\
322.899775789	-237.039273098 \\
};

\addplot [line width = 0.75mm, colorbanana, forget plot]
table [row sep=\\]{%
332.206618804	-293.225737635 \\
328.4001713	-268.9145508 \\
324.374242822	-255.296880791 \\
322.775876973	-248.130054877 \\
322.121718054	-245.729930565 \\
321.01504309	-243.852384175 \\
320.932924457	-243.655404706 \\
320.887614466	-243.552442727 \\
320.981818234	-243.768080527 \\
321.145460351	-243.831079112 \\
};

\addplot [only marks,line width=0.75pt,mark size=3.5pt,mark=*,mark options={solid}, color1]
table [row sep=\\]{%
	320	-240 \\
};

\end{axis}
\end{tikzpicture}
	\includegraphics[width=0.235\textwidth]{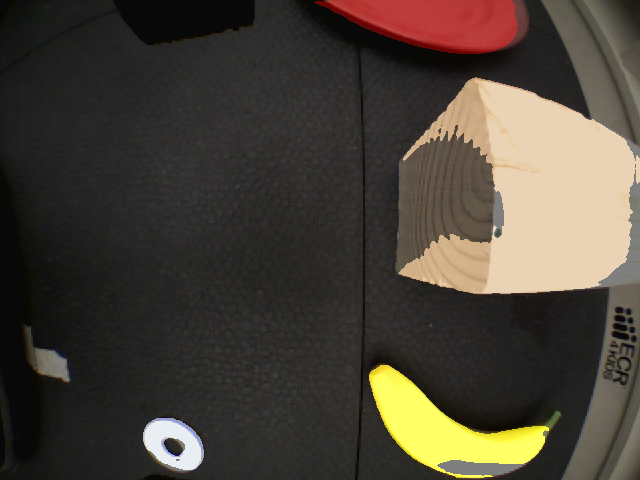}
\begin{tikzpicture}

\definecolor{color1}{rgb}{0.3,1,0.3}
\definecolor{colorbanana}{rgb}{1,0.9,0}
\definecolor{colorwood}{rgb}{0.86,0.71,0.52}
\definecolor{colorwasher}{rgb}{0.86,0.89,0.92}
\definecolor{colorwasher}{rgb}{0.72,0.78,0.84}
\definecolor{colorplate}{rgb}{0.6,0,0}

\begin{axis}[
ticks = none,
tick align=outside,
tick pos=left,
width=5.7cm,
height=4.6675cm,
x grid style={lightgray!92.026143790849673!black},
xmajorgrids,
xmin=0, xmax=640,
xtick={0,160,320,480,640},
y grid style={lightgray!92.026143790849673!black},
grid style={lightgray!50},
legend style={at={(0.02,0.02)},anchor=south west},
legend cell align=left,
xticklabels={},
yticklabels={},
ymin=-480, ymax=0,
ytick={0,-120,-240,-360,-480},
ymajorgrids,
/tikz/inner sep to outer sep/.style={inner sep=0pt, outer sep=.3333em},
x tick label style=inner sep to outer sep,
x label style=inner sep to outer sep,
y label style=inner sep to outer sep,
]
\addplot [line width = 0.75mm, colorwood, forget plot]
table [row sep=\\]{%
537.267631863	-182.948243129 \\
371.736280036	-216.541531826 \\
298.058481953	-231.468981363 \\
267.704534036	-229.739297783 \\
252.181805243	-226.351278861 \\
242.713559851	-224.102191658 \\
238.569975815	-224.022475405 \\
237.073471046	-224.501798892 \\
237.234116736	-224.498889795 \\
237.022677229	-224.531667225 \\
237.254641875	-224.485443845 \\
237.259496955	-224.531467781 \\
237.287968432	-224.482974148 \\
237.17226264	-224.520428214 \\
237.261109654	-224.500128362 \\
237.182423722	-224.479813015 \\
237.310754307	-224.480820471 \\
237.250105895	-224.451271263 \\
237.276884598	-224.418021727 \\
237.471492342	-224.390433265 \\
};

\addplot [line width = 0.75mm, colorwasher, forget plot]
table [row sep=\\]{%
173.117390835	-442.101892298 \\
182.703718226	-367.458062868 \\
197.28496009	-309.384181736 \\
204.961468635	-282.995561846 \\
208.968159312	-269.389108199 \\
210.8992999	-263.956453303 \\
211.72531043	-262.377300266 \\
211.684128997	-262.379084177 \\
212.010497635	-262.162604416 \\
212.309878885	-262.055296693 \\
212.205890401	-262.035055495 \\
212.301568745	-262.069403283 \\
212.271410357	-262.058885268 \\
212.266738275	-262.071896959 \\
212.273477975	-262.039950336 \\
212.416708956	-262.210350094 \\
212.473137019	-262.344104422 \\
212.686240261	-262.301338536 \\
212.979780471	-262.588600061 \\
212.928708747	-262.711643155 \\
};

\addplot [line width = 0.75mm, colorplate, forget plot]
table [row sep=\\]{%
453.282637996	-21.0052323466 \\
341.119398024	-73.4714153079 \\
292.114921065	-116.979828824 \\
248.648092919	-148.030063217 \\
234.728531	-167.044459904 \\
226.997797415	-188.354343451 \\
222.796895944	-195.258883911 \\
220.4092712	-199.918255304 \\
219.421940552	-203.884289528 \\
218.062058771	-207.364219677 \\
217.061548612	-210.940787665 \\
216.824148007	-213.650741781 \\
216.250495478	-213.46158186 \\
216.024976557	-214.981976155 \\
215.978700142	-216.890887013 \\
};

\addplot [line width = 0.75mm, colorbanana, forget plot]
table [row sep=\\]{%
446.611999979	-424.516606608 \\
353.144507876	-352.010937199 \\
291.094974992	-292.84093746 \\
260.505912513	-263.860732053 \\
247.587955859	-250.607165262 \\
243.180160227	-242.917478664 \\
241.60319184	-239.482658511 \\
241.562041894	-239.40085425 \\
241.516481911	-239.391619235 \\
241.547106346	-239.389334937 \\
241.548479371	-239.409767706 \\
241.644606445	-239.401750747 \\
241.571049241	-239.401579262 \\
241.49008406	-239.366620698 \\
241.61837812	-239.41512723 \\
241.531648913	-239.286088179 \\
241.588419204	-239.247548487 \\
241.586293772	-239.247245722 \\
241.602272023	-239.249039001 \\
241.547093923	-239.256663143 \\
};

\addplot [only marks,line width=0.75pt,mark size=3.5pt,mark=*,mark options={solid}, color1]
table [row sep=\\]{%
	220	-240 \\
};

\end{axis}
\end{tikzpicture}
	\caption{
		\textbf{Visual Servoing using Learned Parameters}.
		Initial view with segmented objects (left) and visual servo trajectories centering on each object (right). While objects are identically placed for the $\mathbf{H}_{\text{base}}$ (top) and $\mathbf{H}_{\text{base grasp}}$ (bottom) experiments, each configuration has learned the correct scale of actuation to center on objects from its own visual perspective.
		Note that in the $\mathbf{H}_{\text{base}}$ view, the wood block starts very close to $s^*$ (green dot).
	}
	\label{fig:vs_base}
\end{figure}
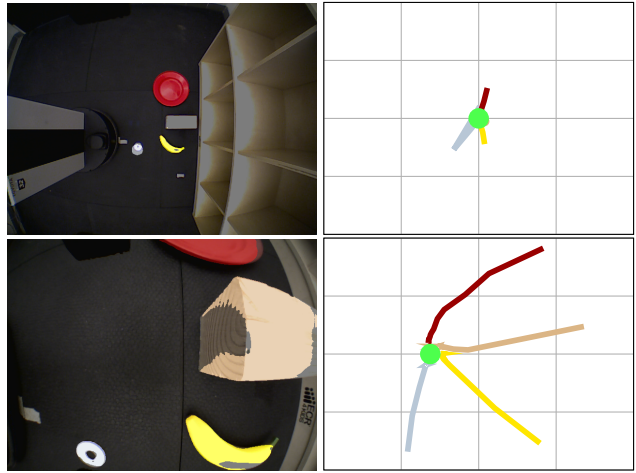

\vspace{5pt}
\noindent\textbf{$\mathbf{H}_{\text{base}}$ Results}
We show the step response of all $\widehat{\mathbf{J}_{\mathbf{s}}^+}$ configurations in Table~\ref{table:lresults} by performing experiments centering the camera on objects placed at various viewpoints within each configuration's starting pose.
In Figure~\ref{fig:vs_base}, both $\mathbf{H}_{\text{base}}$ and $\mathbf{H}_{\text{base grasp}}$ exhibit a stable response. 
Our motivation to learn two base configurations is the increase in $s_{x,y}$ sensitivity to base motion as an object's depth decreases. 
$\mathbf{H}_{\text{base}}$ operates with the camera raised high above objects, while $\mathbf{H}_{\text{base grasp}}$ operates with the camera directly above objects to position for grasping.
Thus, $\mathbf{H}_{\text{base}}$ requires more movement than $\mathbf{H}_{\text{base grasp}}$ for the same changes in $s_{x,y}$.
This difference is apparent in Table~\ref{table:lresults} from $\mathbf{H}_{\text{base}}$ learning greater $\frac{\partial q_{\text{base}}}{\partial s}$ values and in Figure~\ref{fig:vs_base} from $\mathbf{H}_{\text{base}}$'s smaller $s_{x,y}$ distribution for identical object distances.

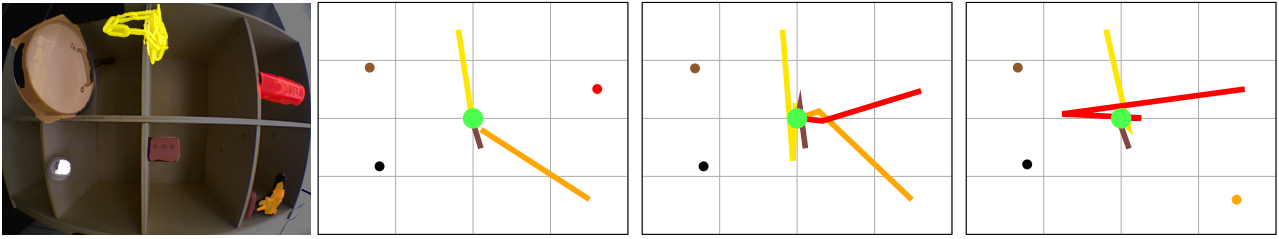
\begin{figure*}
	\centering
	\includegraphics[width=0.235\textwidth]{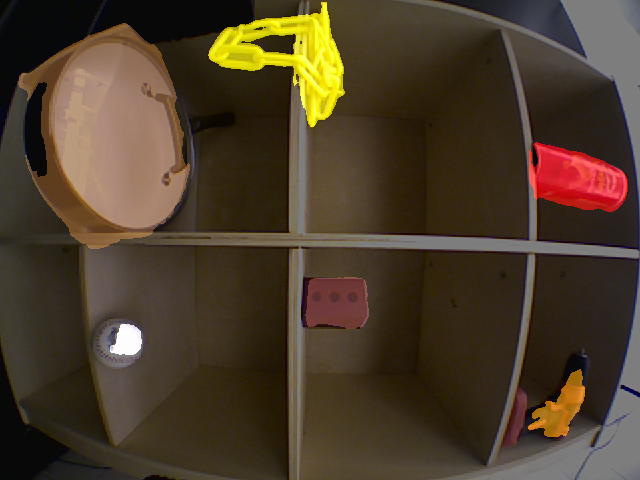}
\begin{tikzpicture}

\definecolor{color0}{rgb}{0.12156862745098,0.466666666666667,0.705882352941177}
\definecolor{color0}{rgb}{0.00000,0.15290,0.29800}

\definecolor{color1}{rgb}{0.85,0.3,0.3}
\definecolor{color1}{rgb}{0.3,1,0.3}
\definecolor{color2}{rgb}{0.93,0.93,0.15}

\definecolor{colorchain}{rgb}{0.9,0.9,0}
\definecolor{colorchain}{rgb}{1.00000,0.79610,0.01961}
\definecolor{colorchain}{rgb}{1.00000,0.9,0}
\definecolor{colortuna}{rgb}{0,0,1}

\definecolor{colorpringles}{rgb}{1,0,0}
\definecolor{colordrill}{rgb}{1,0.65,0}
\definecolor{colorbrick}{rgb}{0.51, 0.28,0.25}
\definecolor{colorbaseball}{rgb}{0,0,0}
\definecolor{colorpan}{rgb}{0.56,0.35,0.16}

\begin{axis}[
ticks = none,
tick align=outside,
tick pos=left,
width=5.7cm,
height=4.6675cm,
x grid style={lightgray!92.026143790849673!black},
xmajorgrids,
xmin=0, xmax=640,
xtick={0,160,320,480,640},
y grid style={lightgray!92.026143790849673!black},
grid style={lightgray!50},
legend style={at={(0.02,0.02)},anchor=south west},
legend cell align=left,
xticklabels={},
yticklabels={},
ymin=-480, ymax=0,
ytick={0,-120,-240,-360,-480},
ymajorgrids,
/tikz/inner sep to outer sep/.style={inner sep=0pt, outer sep=.3333em},
x tick label style=inner sep to outer sep,
x label style=inner sep to outer sep,
y label style=inner sep to outer sep,
]

\addplot [line width = 0.75mm, colorchain, forget plot]
table [row sep=\\]{%
289.565272075	-56.2150612848 \\
320.30854715	-253.936121418 \\
320.12973783	-242.466213494 \\
320.374352557	-237.897181506 \\
320.383856732	-240.096766358 \\
};

\addplot [line width = 0.75mm, colordrill, forget plot]
table [row sep=\\]{%
560.443764515	-407.743271176 \\
336.550041882	-263.029865857 \\
};

\addplot [only marks,line width=0.75pt,mark size=1.5pt,mark=*,mark options={solid}, colorpringles]
table [row sep=\\]{%
576.64354506	-179.126841239 \\
};

\addplot [line width = 0.75mm, colorbrick, forget plot]
table [row sep=\\]{%
336.320821805	-301.946860607 \\
316.350090939	-237.159597073 \\
319.770674204	-239.536390959 \\
319.916130377	-240.086390411 \\
};

\addplot [only marks,line width=0.75pt,mark size=1.5pt,mark=*,mark options={solid}, colorbaseball]
table [row sep=\\]{%
	126.871077015	-339.19313943 \\
};

\addplot [only marks,line width=0.75pt,mark size=1.5pt,mark=*,mark options={solid}, colorpan]
table [row sep=\\]{%
106.504310372	-134.842545513 \\
};

\addplot [only marks,line width=0.75pt,mark size=3.5pt,mark=*,mark options={solid}, color1]
table [row sep=\\]{%
	320	-240 \\
};

\end{axis}
\end{tikzpicture}
\begin{tikzpicture}

\definecolor{color0}{rgb}{0.12156862745098,0.466666666666667,0.705882352941177}
\definecolor{color0}{rgb}{0.00000,0.15290,0.29800}

\definecolor{color1}{rgb}{0.85,0.3,0.3}
\definecolor{color1}{rgb}{0.3,1,0.3}
\definecolor{color2}{rgb}{0.93,0.93,0.15}

\definecolor{colorchain}{rgb}{0.9,0.9,0}
\definecolor{colorchain}{rgb}{1.00000,0.9,0}
\definecolor{colortuna}{rgb}{0,0,1}

\definecolor{colorpringles}{rgb}{1,0,0}
\definecolor{colordrill}{rgb}{1,0.65,0}
\definecolor{colorbrick}{rgb}{0.51, 0.28,0.25}
\definecolor{colorbaseball}{rgb}{0,0,0}
\definecolor{colorpan}{rgb}{0.56,0.35,0.16}

\begin{axis}[
ticks = none,
tick align=outside,
tick pos=left,
width=5.7cm,
height=4.6675cm,
x grid style={lightgray!92.026143790849673!black},
xmajorgrids,
xmin=0, xmax=640,
xtick={0,160,320,480,640},
y grid style={lightgray!92.026143790849673!black},
grid style={lightgray!50},
legend style={at={(0.02,0.02)},anchor=south west},
legend cell align=left,
xticklabels={},
yticklabels={},
ymin=-480, ymax=0,
ytick={0,-120,-240,-360,-480},
ymajorgrids,
/tikz/inner sep to outer sep/.style={inner sep=0pt, outer sep=.3333em},
x tick label style=inner sep to outer sep,
x label style=inner sep to outer sep,
y label style=inner sep to outer sep,
]

\addplot [line width = 0.75mm, colorchain, forget plot]
table [row sep=\\]{%
	290.106786835	-56.501517711 \\
	311.265567489	-327.107066272 \\
	317.226955543	-208.196609338 \\
	318.918637579	-246.078924272 \\
	319.529192482	-243.870944986 \\
	320.157659582	-238.1488654 \\
	320.332104198	-236.829226005 \\
	320.036286014	-240.335568406 \\
};

\addplot [line width = 0.75mm, colordrill, forget plot]
table [row sep=\\]{%
	556.977831916	-408.277361143 \\
	365.129731693	-225.058415447 \\
	333.760366452	-236.26665405 \\
	324.767753744	-240.479382759 \\
	322.374654729	-240.682418408 \\
	320.772551412	-239.998523622 \\
};

\addplot [line width = 0.75mm, colorpringles, forget plot]
table [row sep=\\]{%
	576.194025778	-182.914189557 \\
	372.65636803	-245.194477475 \\
	340.76923901	-241.715333772 \\
	322.377483683	-239.855245387 \\
	320.738545067	-239.700408902 \\
	320.478189867	-240.012117501 \\
};

\addplot [line width = 0.75mm, colorbrick, forget plot]
table [row sep=\\]{%
	337.242543496	-301.782533978 \\
	325.792057584	-215.208620359 \\
	320.798690349	-240.625068548 \\
};

\addplot [only marks,line width=0.75pt,mark size=1.5pt,mark=*,mark options={solid}, colorbaseball]
table [row sep=\\]{%
	126.871077015	-339.19313943 \\
};

\addplot [only marks,line width=0.75pt,mark size=1.5pt,mark=*,mark options={solid}, colorpan]
table [row sep=\\]{%
	109.142314455	-136.436000148 \\
};

\addplot [only marks,line width=0.75pt,mark size=3.5pt,mark=*,mark options={solid}, color1]
table [row sep=\\]{%
	320	-240 \\
};

\end{axis}
\end{tikzpicture}
\begin{tikzpicture}

\definecolor{color0}{rgb}{0.12156862745098,0.466666666666667,0.705882352941177}
\definecolor{color0}{rgb}{0.00000,0.15290,0.29800}

\definecolor{color1}{rgb}{0.85,0.3,0.3}
\definecolor{color1}{rgb}{0.3,1,0.3}
\definecolor{color2}{rgb}{0.93,0.93,0.15}

\definecolor{colorchain}{rgb}{0.9,0.9,0}
\definecolor{colorchain}{rgb}{1.00000,0.79610,0.01961}
\definecolor{colorchain}{rgb}{1.00000,0.9,0}
\definecolor{colortuna}{rgb}{0,0,1}

\definecolor{colorpringles}{rgb}{1,0,0}
\definecolor{colordrill}{rgb}{1,0.65,0}
\definecolor{colorbrick}{rgb}{0.51, 0.28,0.25}
\definecolor{colorbaseball}{rgb}{0,0,0}
\definecolor{colorpan}{rgb}{0.56,0.35,0.16}

\begin{axis}[
ticks = none,
tick align=outside,
tick pos=left,
width=5.7cm,
height=4.6675cm,
x grid style={lightgray!92.026143790849673!black},
xmajorgrids,
xmin=0, xmax=640,
xtick={0,160,320,480,640},
y grid style={lightgray!92.026143790849673!black},
grid style={lightgray!50},
legend style={at={(0.02,0.02)},anchor=south west},
legend cell align=left,
xticklabels={},
yticklabels={},
ymin=-480, ymax=0,
ytick={0,-120,-240,-360,-480},
ymajorgrids,
/tikz/inner sep to outer sep/.style={inner sep=0pt, outer sep=.3333em},
x tick label style=inner sep to outer sep,
x label style=inner sep to outer sep,
y label style=inner sep to outer sep,
]

\addplot [line width = 0.75mm, colorchain, forget plot]
table [row sep=\\]{%
288.946582636	-56.4069892131 \\
335.068529606	-258.725330208 \\
314.683446163	-240.691342743 \\
320.449922381	-238.859326081 \\
320.167534994	-240.409433207 \\
};

\addplot [only marks,line width=0.75pt,mark size=1.5pt,mark=*,mark options={solid}, colordrill]
table [row sep=\\]{%
558.614362444	-408.18625392 \\
};

\addplot [line width = 0.75mm, colorpringles, forget plot]
table [row sep=\\]{%
575.334882997	-179.396672729 \\
198.197908099	-230.734185592 \\
361.425036169	-238.998057441 \\
310.569823662	-239.651822214 \\
319.690541266	-240.133900816 \\
};

\addplot [line width = 0.75mm, colorbrick, forget plot]
table [row sep=\\]{%
336.332340768	-302.299435492 \\
312.576163003	-236.27209398 \\
321.579854611	-239.345554688 \\
320.602823937	-239.732833779 \\
320.292532132	-240.001858464 \\
};

\addplot [only marks,line width=0.75pt,mark size=1.5pt,mark=*,mark options={solid}, colorbaseball]
table [row sep=\\]{%
125.923849773	-335.13448748 \\
};

\addplot [only marks,line width=0.75pt,mark size=1.5pt,mark=*,mark options={solid}, colorpan]
table [row sep=\\]{%
106.504310372	-134.842545513 \\
};

\addplot [only marks,line width=0.75pt,mark size=3.5pt,mark=*,mark options={solid}, color1]
table [row sep=\\]{%
	320	-240 \\
};

\end{axis}
\end{tikzpicture}
	\caption{Initial view of objects and visual servo trajectories using $\mathbf{H}_{\text{arm lift}}$ (center left),  $\mathbf{H}_{\text{arm wrist}}$ (center right), and $\mathbf{H}_{\text{arm both}}$ (right).}
	\label{fig:vs_arm}
\end{figure*}

\vspace{5pt}
\noindent\textbf{$\mathbf{H}_{\text{arm}}$ Results}
We show the step response of all arm-based VOS-VS configurations in Figure~\ref{fig:vs_arm}.
Each configuration uses the same objects and starting pose.
Although each configuration segments the pan and baseball, $s^*$ is not reachable for these objects within any of the configured actuator spaces; $\mathbf{H}_{\text{arm wrist}}$ is the only configuration to center on all four of the other objects.
The overactuated $\mathbf{H}_{\text{arm both}}$ has the most overshoot, while $\mathbf{H}_{\text{arm lift}}$ has the most limited range of camera positions but essentially deadbeat control.


\begin{figure}
	\centering
	\includegraphics[width=0.235\textwidth]{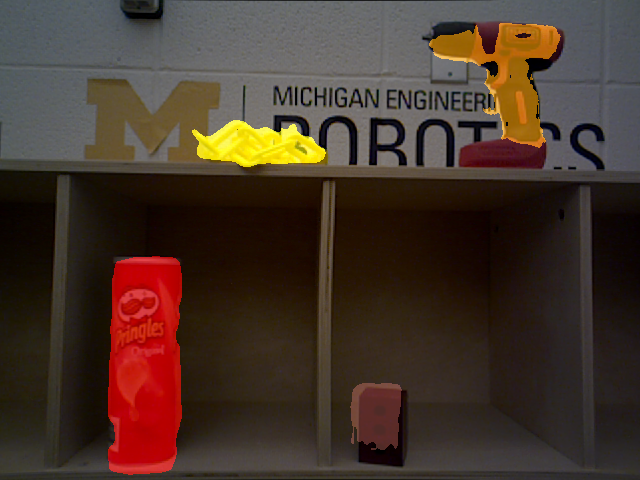}
\begin{tikzpicture}

\definecolor{color0}{rgb}{0.12156862745098,0.466666666666667,0.705882352941177}
\definecolor{color0}{rgb}{0.00000,0.15290,0.29800}

\definecolor{color1}{rgb}{0.85,0.3,0.3}
\definecolor{color1}{rgb}{0.3,1,0.3}
\definecolor{color2}{rgb}{0.93,0.93,0.15}

\definecolor{colorchain}{rgb}{0.9,0.9,0}
\definecolor{colorchain}{rgb}{1.00000,0.79610,0.01961}
\definecolor{colorchain}{rgb}{1.00000,0.9,0}
\definecolor{colortuna}{rgb}{0,0,1}

\definecolor{colorpringles}{rgb}{1,0,0}
\definecolor{colordrill}{rgb}{1,0.65,0}
\definecolor{colorbrick}{rgb}{0.51, 0.28,0.25}
\definecolor{colorbaseball}{rgb}{0,0,0}
\definecolor{colorpan}{rgb}{0.56,0.35,0.16}

\begin{axis}[
ticks = none,
tick align=outside,
tick pos=left,
width=5.7cm,
height=4.6675cm,
x grid style={lightgray!92.026143790849673!black},
xmajorgrids,
xmin=0, xmax=640,
xtick={0,160,320,480,640},
y grid style={lightgray!92.026143790849673!black},
grid style={lightgray!50},
legend style={at={(0.02,0.02)},anchor=south west},
legend cell align=left,
xticklabels={},
yticklabels={},
ymin=-480, ymax=0,
ytick={0,-120,-240,-360,-480},
ymajorgrids,
/tikz/inner sep to outer sep/.style={inner sep=0pt, outer sep=.3333em},
x tick label style=inner sep to outer sep,
x label style=inner sep to outer sep,
y label style=inner sep to outer sep,
]

\addplot [line width = 0.75mm, colorbrick, forget plot]
table [row sep=\\]{%
	383.878423676	-419.076822081 \\
	333.40940743	-295.143219976 \\
	319.789133937	-252.20460604 \\
	314.846249157	-227.962952814 \\
	318.058821316	-230.079525901 \\
	321.478069381	-238.046121859 \\
	321.134083913	-242.442982423 \\
	319.83610451	-241.279153955 \\
	319.732941025	-240.642992778 \\
	320.102375813	-240.458959334 \\
};

\addplot [line width = 0.75mm, colorchain, forget plot]
table [row sep=\\]{%
263.777129564	-133.718250589 \\
305.366687744	-219.089268711 \\
318.954491357	-252.490429848 \\
321.369138617	-248.65026253 \\
320.86700052	-241.841922724 \\
320.214625472	-237.233240332 \\
318.72447824	-237.7633167 \\
319.644903731	-240.404267324 \\
320.196507252	-241.062274363 \\
320.924188551	-240.440399282 \\
};

\addplot [line width = 0.75mm, colordrill, forget plot]
table [row sep=\\]{%
524.288366281	-76.5772950232 \\
326.640689836	-224.851802583 \\
319.207698721	-240.503420889 \\
};

\addplot [line width = 0.75mm, colorpringles, forget plot]
table [row sep=\\]{%
163.148217021	-336.343016156 \\
281.246041981	-290.670358657 \\
312.829072945	-257.113982744 \\
327.590715506	-236.798561371 \\
325.848844949	-232.708967554 \\
319.245614006	-238.394170473 \\
318.502834214	-240.373684203 \\
319.912353989	-240.296077673 \\
320.021294692	-240.025407938 \\
320.071345633	-240.16461122 \\
};

\addplot [only marks,line width=0.75pt,mark size=3.5pt,mark=*,mark options={solid}, color1]
table [row sep=\\]{%
	320	-240 \\
};

\end{axis}
\end{tikzpicture}
	\caption{Initial view and visual servo trajectories using $\mathbf{H}_{\text{head}}$.}
	\label{fig:vs_head}
\end{figure}
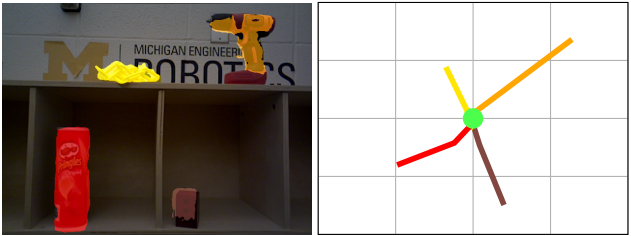

\vspace{5pt}
\noindent\textbf{$\mathbf{H}_{\text{head}}$ Results}
Finally, we show the step response of $\mathbf{H}_{\text{head}}$ in Figure~\ref{fig:vs_head}.
$\mathbf{H}_{\text{head}}$ is the only configuration that uses HSR's 2-DOF head gimbal and camera, and it exhibits a smooth step response over the entire image.
Remarkably, even though $\mathbf{H}_{\text{head}}$ uses the head camera, it still uses the same OSVOS parameters $\mathbf{W}$ that are learned on grasp camera images; this further demonstrates the general applicability of VOS-VS in regards to needing no camera calibration.  

\subsection{Consecutive Mobile Robot Trials}
\label{sec:vsde}

We perform an experiment consisting of a consecutive set of mobile trials that simultaneously test VOS-VS and VOS-DE.
Each trial consists of three unique YCB objects placed at different heights: one on the blue bin 0.25~\textrm{m} above the ground, one on the green bin 0.125~\textrm{m} above the ground, and one directly on the ground (see bin configuration in Figure~\ref{fig:stairs}).
The trial configurations and corresponding results are provided in Table~\ref{table:expRealWorld}. 
VOS-VS is considered a success (``X'') if HSR locates and centers on the object for depth estimation.
VOS-DE is considered a success if HSR achieves $z_{\text{camera, grasp}}$ \eqref{eq:grasp} such that HSR can close its grippers on the object without hitting the underlying surface and $z_{\text{camera}}$ does not move past the top surface of the object.

\setlength{\tabcolsep}{9.5pt}
\begin{table}
	\centering
	\caption{\textbf{Consecutive Mobile Robot Trial Results.}
		All results are from a single consecutive set of mobile HSR trials. Across all of the challenge objects, VOS-VS has a 83\% success rate. Except for one VOS-DE trial, the food objects were a complete success.
	}
	\footnotesize
	\begin{tabular}{ l | l | l | c | c}
		\hline 
		\multicolumn{1}{c|}{} & \multicolumn{1}{c|}{Object} & \multicolumn{1}{c|}{Support} & \multicolumn{2}{c}{Success} \\
		\cline{4-5}
		\multicolumn{1}{c|}{Item} & \multicolumn{1}{c|}{Category} & \multicolumn{1}{c|}{Height (\textrm{m})} & VS & DE \\
		\hline
		\rowcolor{rowgray}	Chips Can	&	Food	&	0.25	&	X	&	X	\\
		Potted Meat	&	Food	&	0.125	&	X	&	X	\\
		\rowcolor{rowgray}	Plastic Banana	&	Food	&	Ground	&	X	&	X	\\
		\hline									
		Box of Sugar	&	Food	&	0.25	&	X	&	X	\\
		\rowcolor{rowgray}	Tuna	&	Food	&	0.125	&	X	&		\\
		Gelatin	&	Food	&	Ground	&	X	&	X	\\
		\hline									
		\rowcolor{rowgray}	Mug	&	Kitchen	&	0.25	&	X	&	X	\\
		Softscrub	&	Kitchen	&	0.125	&		&	N/A	\\
		\rowcolor{rowgray}	Skillet with Lid	&	Kitchen	&	Ground	&		&	N/A	\\
		\hline									
		Plate	&	Kitchen	&	0.25	&	X	&	X	\\
		\rowcolor{rowgray}	Spatula	&	Kitchen	&	0.125	&		&	N/A	\\
		Knife	&	Kitchen	&	Ground	&	X	&		\\
		\hline									
		\rowcolor{rowgray}	Power Drill	&	Tool	&	0.25	&	X	&	X	\\
		Marker	&	Tool	&	0.125	&	X &		\\
		\rowcolor{rowgray}	Padlock	&	Tool	&	Ground	&	X	&		\\
		\hline									
		Wood	&	Tool	&	0.25	&	X	&		\\
		\rowcolor{rowgray}	Spring Clamp	&	Tool	&	0.125	&	X	&		\\
		Screwdriver	&	Tool	&	Ground	&	X	&		\\
		\hline									
		\rowcolor{rowgray}	Baseball	&	Shape	&	0.25	&	X	&		\\
		Plastic Chain	&	Shape	&	0.125	&	X	&		\\
		\rowcolor{rowgray}	Washer	&	Shape	&	Ground	&	X	&		\\
		\hline									
		Stacking Cup	&	Shape	&	0.25	&	X	&	X	\\
		\rowcolor{rowgray}	Dice	&	Shape	&	0.125	&		&	N/A	\\
		Foam Brick	&	Shape	&	Ground	&	X	&	X	\\
		\hline
	\end{tabular}
	\label{table:expRealWorld}
\end{table}
\begin{figure}
	\centering
	\includegraphics[width=0.4625\textwidth]{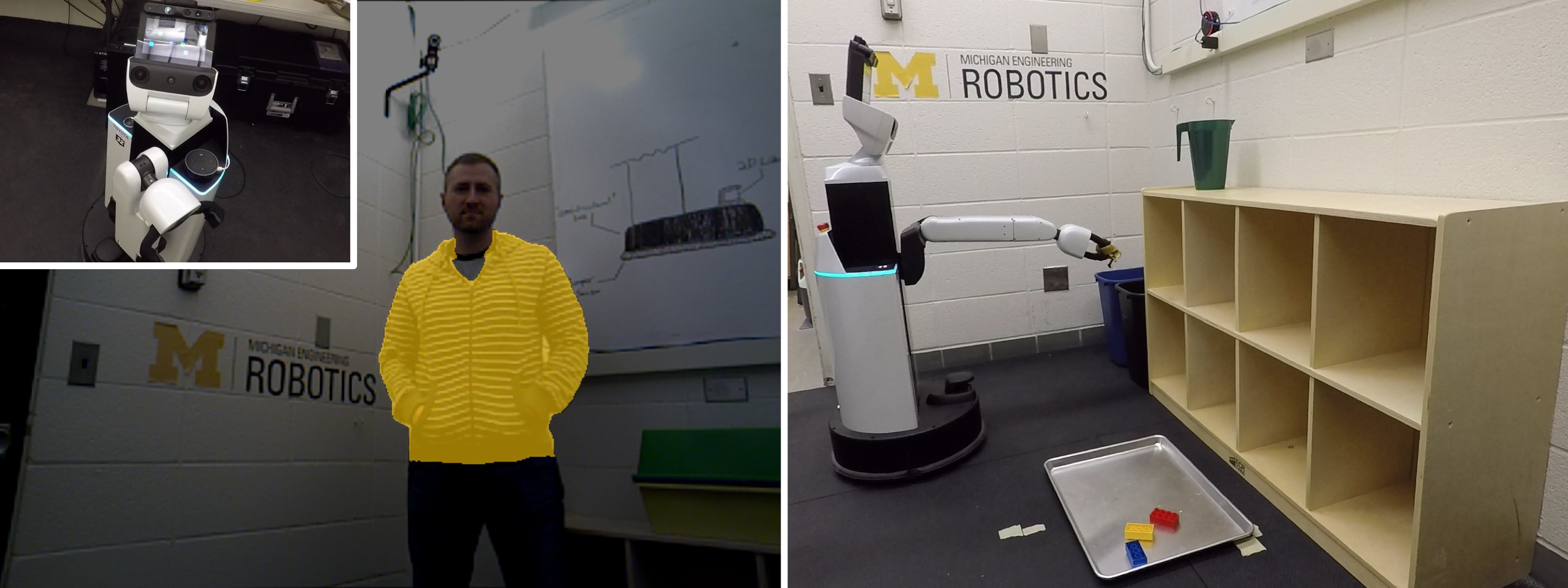}
	\caption{\textbf{Additional Experiments.} Using VOS-VS, HSR is able to track dynamic objects like people in real-time, making VOS-VS a useful tool for human-robot cooperation (left). HSR taking banana peel to garbage for a pick-and-place challenge (right).}
	\label{fig:add}
\end{figure}

Across all 24 objects, VOS-VS has a 83\% success rate.
VOS-DE, which is only applicable when VOS-VS succeeds, has a 50\% success rate.
By category, food objects have the highest success (100\% VOS-VS, 83\% VOS-DE) and kitchen objects have the lowest (50\% VOS-VS, 66\% VOS-DE).
Failures are caused by segmentation errors.
Although VOS-VS can center on a poorly segmented object, VOS-DE fails if there are erratic changes in segmentation area (we provide examples in the Appendix).
Additionally, VOS-DE's margin for success varies between objects (e.g., the smallest margin is the 4~\textrm{mm} thick washer).



\subsection{Additional Experiments}



\noindent\textbf{Pick-and-place Challenges}
We perform additional experiments for our VOS-based methods, 
including our work in the TRI-sponsored HSR challenges.
These challenges consist of timed trials for pick-and-place tasks with randomly scattered, non-YCB objects (e.g., the banana peel in Figure~\ref{fig:add}).
These challenges are a particularly good demonstration of VOS-VS and VOS-Grasp. 
We provide additional figures for these experiments in the Appendix.


%
%
%
%
%

\vspace{10pt}
\noindent\textbf{Dynamic Articulated Objects}
Finally, we perform additional VOS-VS experiments with dynamic articulated objects.
Using $\mathbf{H}_{\text{base}}$, HSR tracks a plastic chain across the room in real-time as we kick it and throw it in a variety of unstructured poses; we can even pick up the chain and use it the guide HSR's movements from the grasp camera.
In addition, by training OSVOS to recognize an article of clothing, HSR reliably tracks a person moving throughout the room using  $\mathbf{H}_{\text{head}}$ (see Figure~\ref{fig:add}).
Experiment videos are available at: \url{https://youtu.be/hlog5FV9RLs}.


\section{Conclusions and Future Work}

We develop a video object segmentation-based approach to visual servo control, depth estimation, and grasping.
Visual servo control is a useful framework for controlling a physical robot system from RGB images, and video object segmentation has seen rampant advances within the computer vision community for densely segmenting unstructured objects in challenging videos.
The success of our segmentation-based approach to visual servo control in mobile robot experiments with real-world objects is a tribute to both of these communities and the initiation of a bridge between them.
Future developments in video object segmentation will improve the robustness of our method and, we expect, lead to other innovations in robotics.

A significant benefit of our segmentation-based framework is that it only requires an RGB camera combined with robot actuation. 
For future work, we are improving RGB-based depth estimation and grasping by comparing images collected from more robot poses, thereby leveraging more information and making our 3D understanding of the target object more complete. 


\vspace{5pt}
\noindent\textbf{Acknowledgment}
Toyota Research Institute (``TRI'') provided funds to assist the authors with their research but this article solely reflects the opinions and conclusions of its authors and not TRI or any other Toyota entity.

\section*{Appendix}
\label{sec:app}

\begin{figure}
	\centering
	\includegraphics[width=0.475\textwidth]{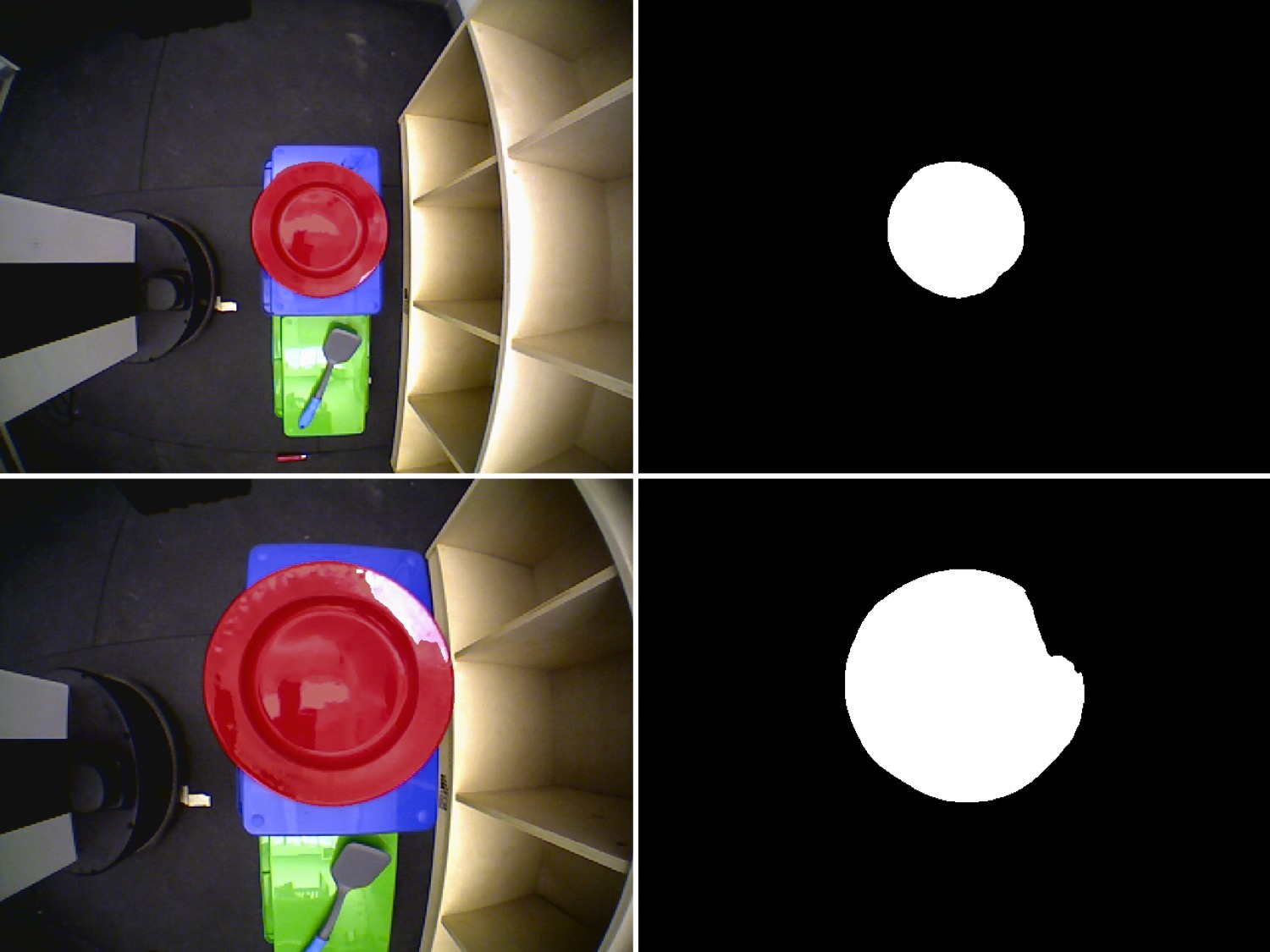}
	\caption{\textbf{Plate Segmentations used for Depth Estimation}. The plate is well-segmented from the higher camera position (top), but has greater spectral reflection as the camera approaches (bottom).
	}
	\label{fig:plate}
\end{figure}

\begin{figure}
	\centering
	\includegraphics[width=0.475\textwidth]{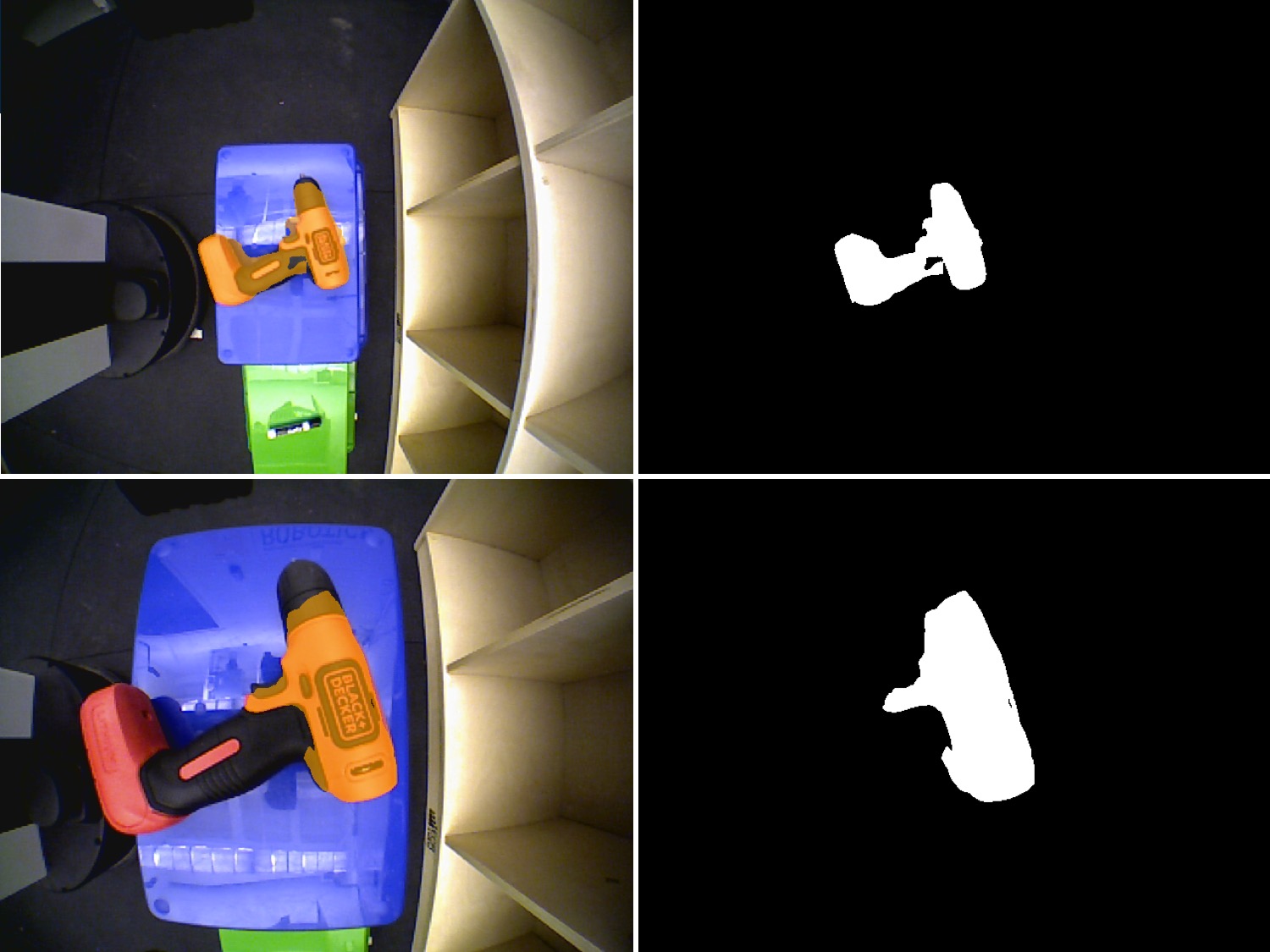}
	\caption{\textbf{Drill Segmentations used for Depth Estimation}. Portions of the drill become unsegmented at the closer view (bottom).
	}
	\label{fig:drill}
\end{figure}
\begin{figure}
	\centering
	\includegraphics[width=0.475\textwidth]{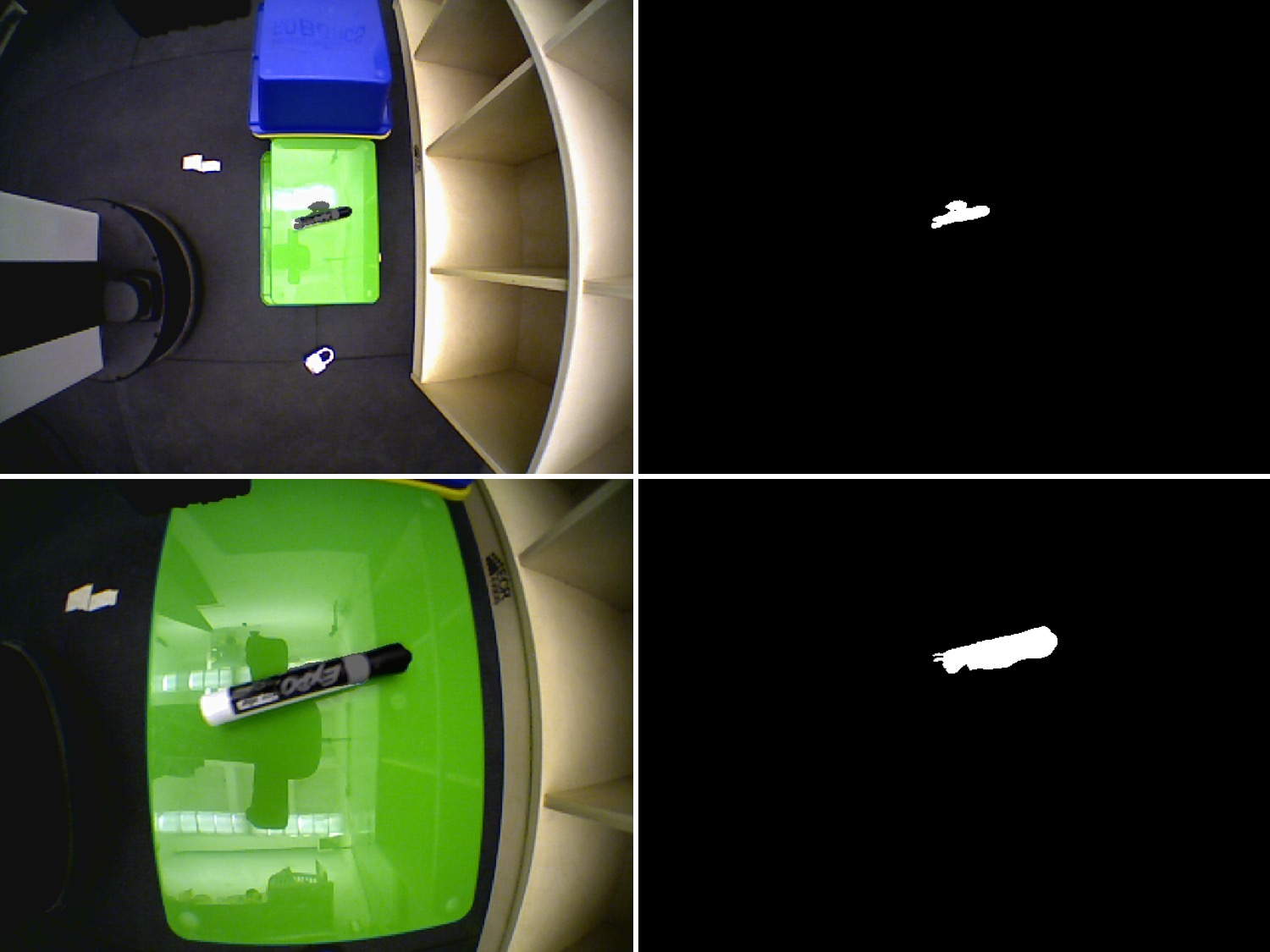}
	\caption{\textbf{Marker Segmentations used for Depth Estimation}. Reflective areas of the background are included as part of the marker segmentation at the higher view (top), then portions of the marker become unsegmented at the closer view (bottom).
	}
	\label{fig:marker}
\end{figure}
\begin{figure}
	\centering
	\includegraphics[width=0.475\textwidth]{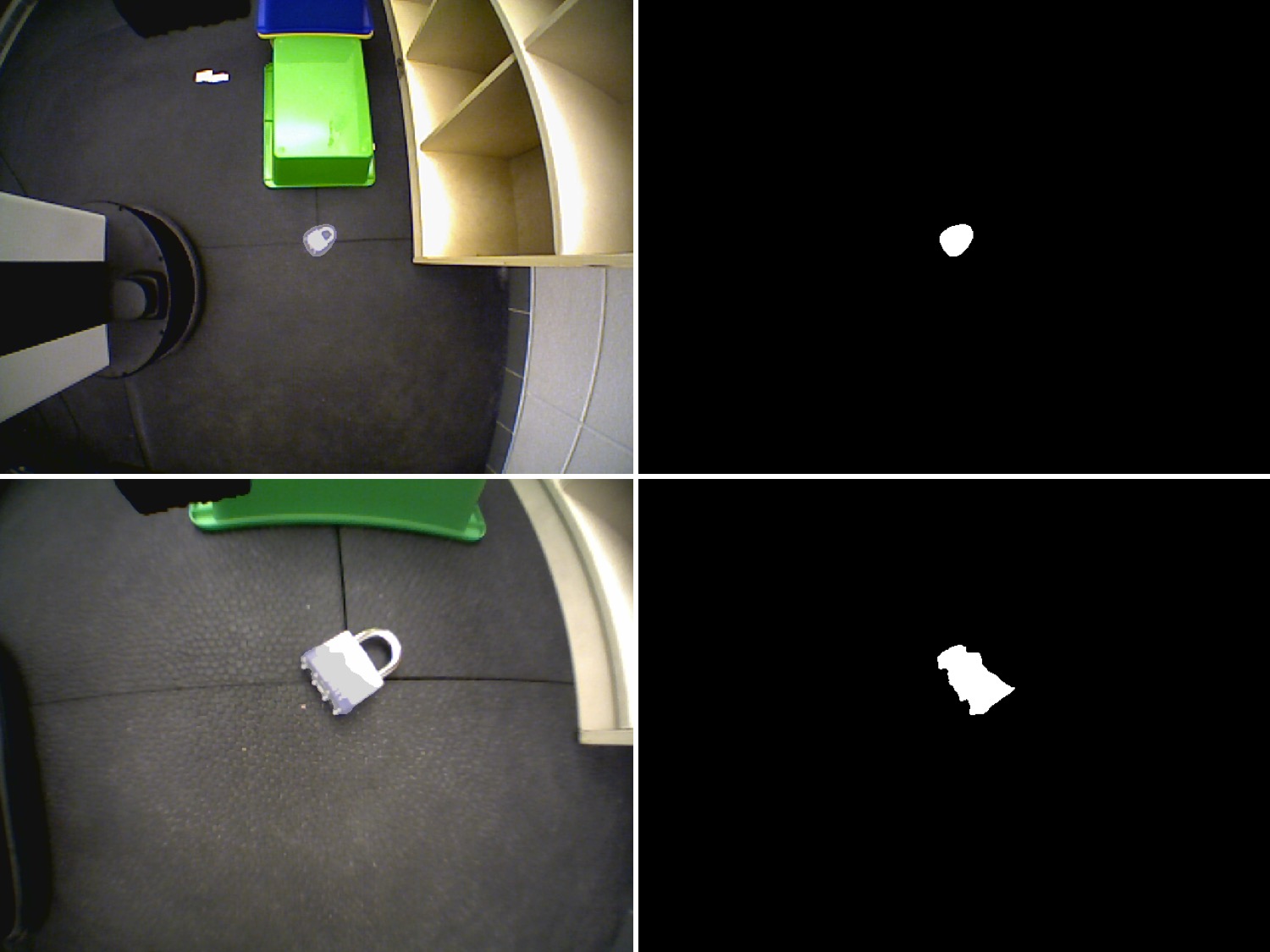}
	\caption{\textbf{Padlock Segmentations used for Depth Estimation}.
		The padlock segmentation goes from including small portions of the background (top) to leaving out large portions of the lock (bottom) as the camera approaches. Segmenting the padlock is difficult due to its small size and specular reflection, and depth estimation of the padlock is difficult due to erroneous changes in segmentation area.
	}
	\label{fig:lock}
\end{figure}

\noindent\textbf{Segmentation Errors} Densely segmenting unstructured objects is a challenging problem, and, despite using state-of-the-art video object segmentation, we have some segmentation errors during our experiments.
Figures~\ref{fig:plate}-\ref{fig:lock} show segmentations used for depth estimation during the consecutive mobile robot trials in Section~\ref{sec:vsde}.
Even with some segmentation errors, VOS-VS centers on all four objects from the high-camera position and VOS-DE successfully estimates the depth of the plate and drill.

\begin{figure*}[h!]
	\centering
	\includegraphics[width=0.975\textwidth]{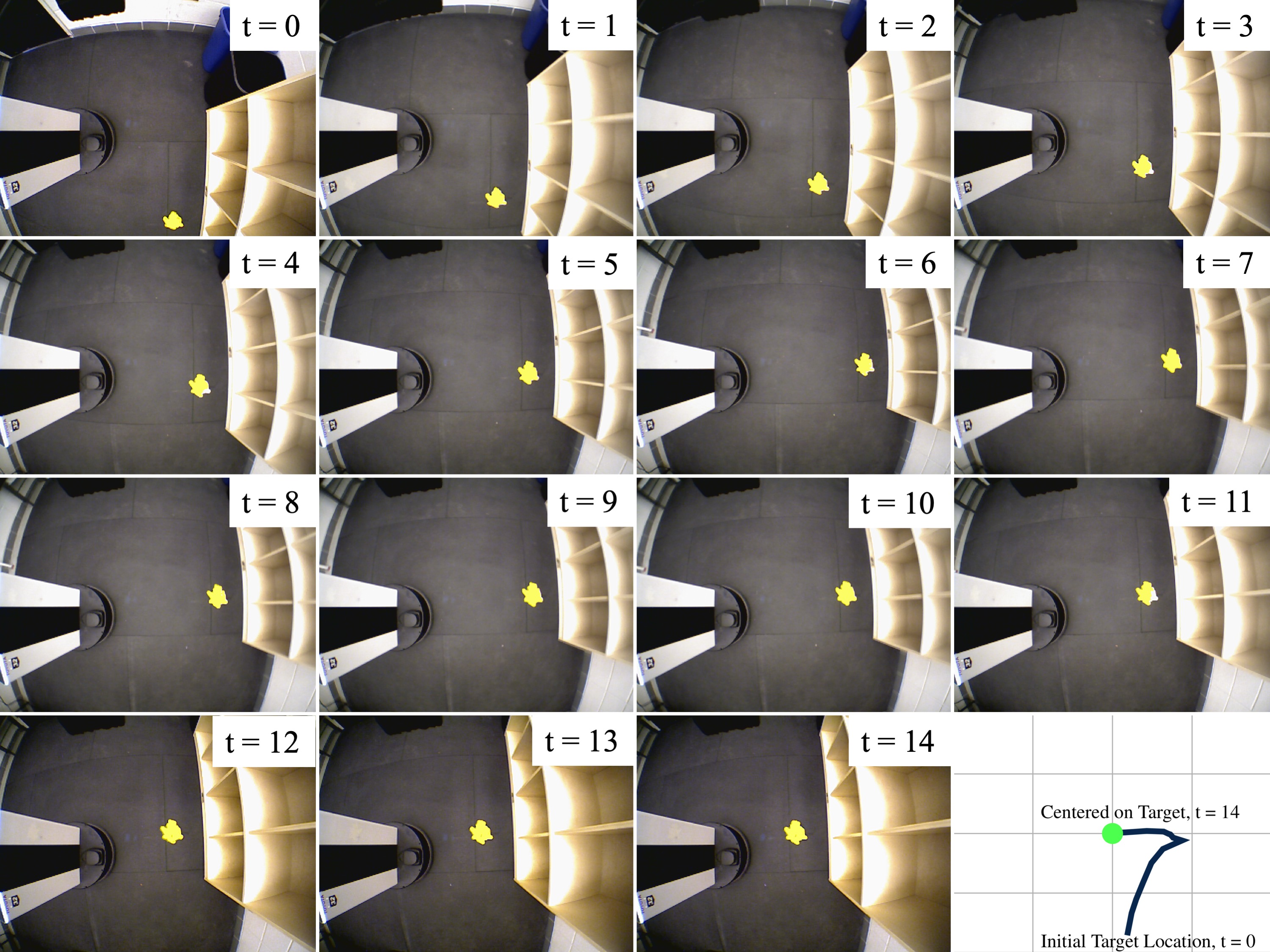}
	\caption{\textbf{Robot Perspective while Learning $\widehat{\mathbf{J}_\mathbf{s}^+}$ for $\mathbf{H}_{\text{base}}$}.  
		Starting with $\widehat{\mathbf{J}_\mathbf{s}^+}_{t=0}$ and offset target location (the yellow chain segmentation), our Hadamard-Broyden update learns the correct visual servoing parameters to center the robot on the target in real-time.
		The target is centered vertically after five updates (t = 5) and horizontally after fourteen (t = 14).
		We show the complete visual servo trajectory of the target object through image space on the bottom right.
		This figure corresponds with the experiment shown in Figures~\ref{fig:vs_init}-\ref{fig:init_param}.
	}
	\label{fig:hb}
\end{figure*}

\vspace{5pt}
\noindent\textbf{Robot's Perspective when Learning VOS-VS} Figure~\ref{fig:hb} shows the step-to-step visual servo transitions from the robot's perspective as it is learning $\widehat{\mathbf{J}_\mathbf{s}^+}$ for $\mathbf{H}_{\text{base}}$ (corresponding to Figures~\ref{fig:vs_init}-\ref{fig:init_param}).

\begin{figure*}
	\centering
	\includegraphics[width=0.975\textwidth]{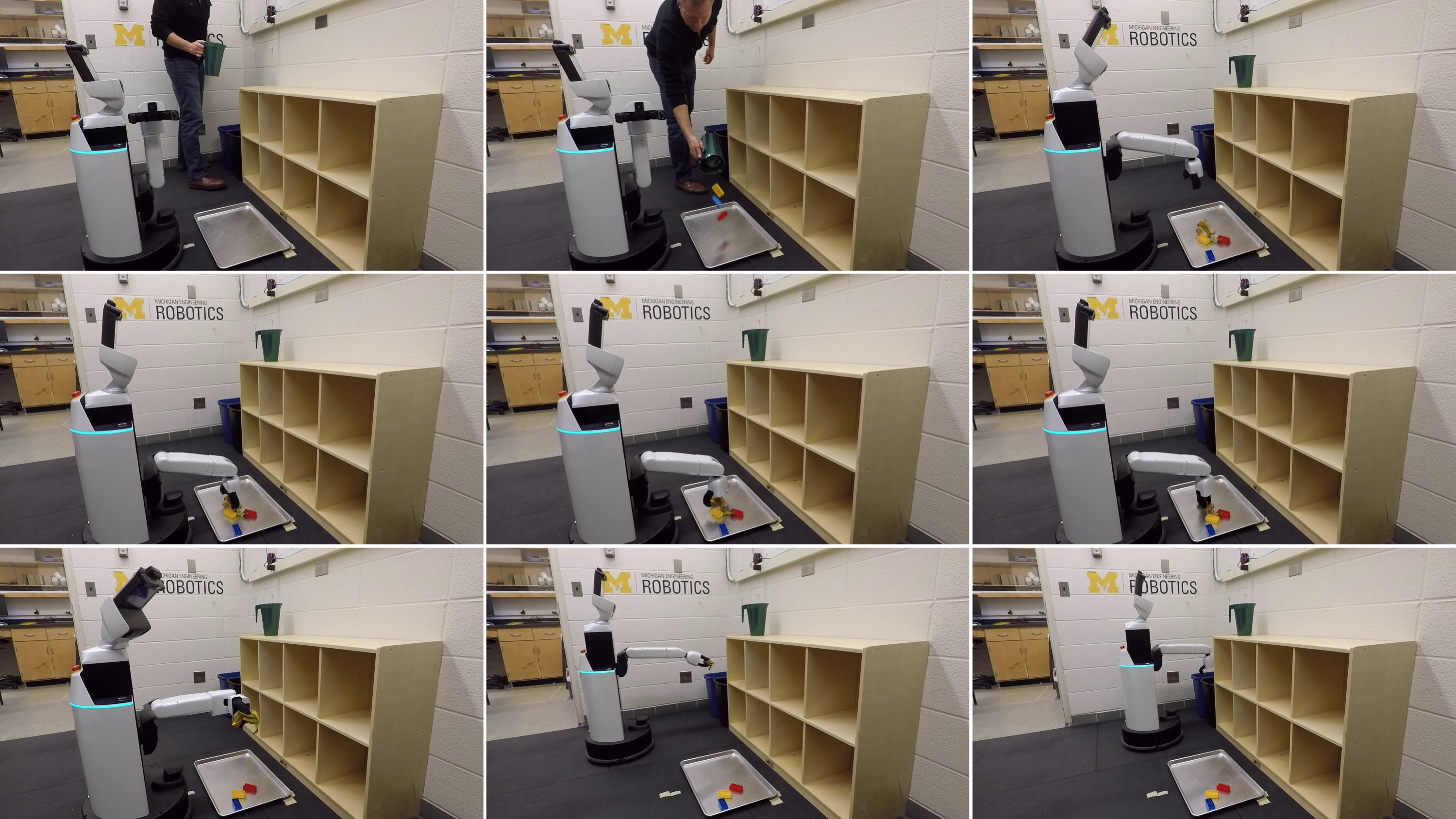}
	\caption{\textbf{HSR using VOS-VS and VOS-Grasp for Pick-and-place}. 
		After a set of HSR challenge objects are randomly poured onto the metal tray, HSR identifies the initial object locations using the downward-facing grasp camera (top row).
		HSR identifies the banana peel as the first target, then centers on the peel amongst the cluttered objects using VOS-VS and then grasps the peel using VOS-Grasp (middle row).
		Finally, HSR performs a visual grasp check away from the other objects and then places the peel in the garbage bin (bottom row). 
		All pick-and-place experiment and trial videos are available at: \url{https://www.youtube.com/playlist?list=PLz52BAn_JPx8nVgP2XfnG_9TCJj0DwC5y}.
	}
	\label{fig:banana}
\end{figure*}

\begin{figure*}
	\centering
	\includegraphics[width=0.975\textwidth]{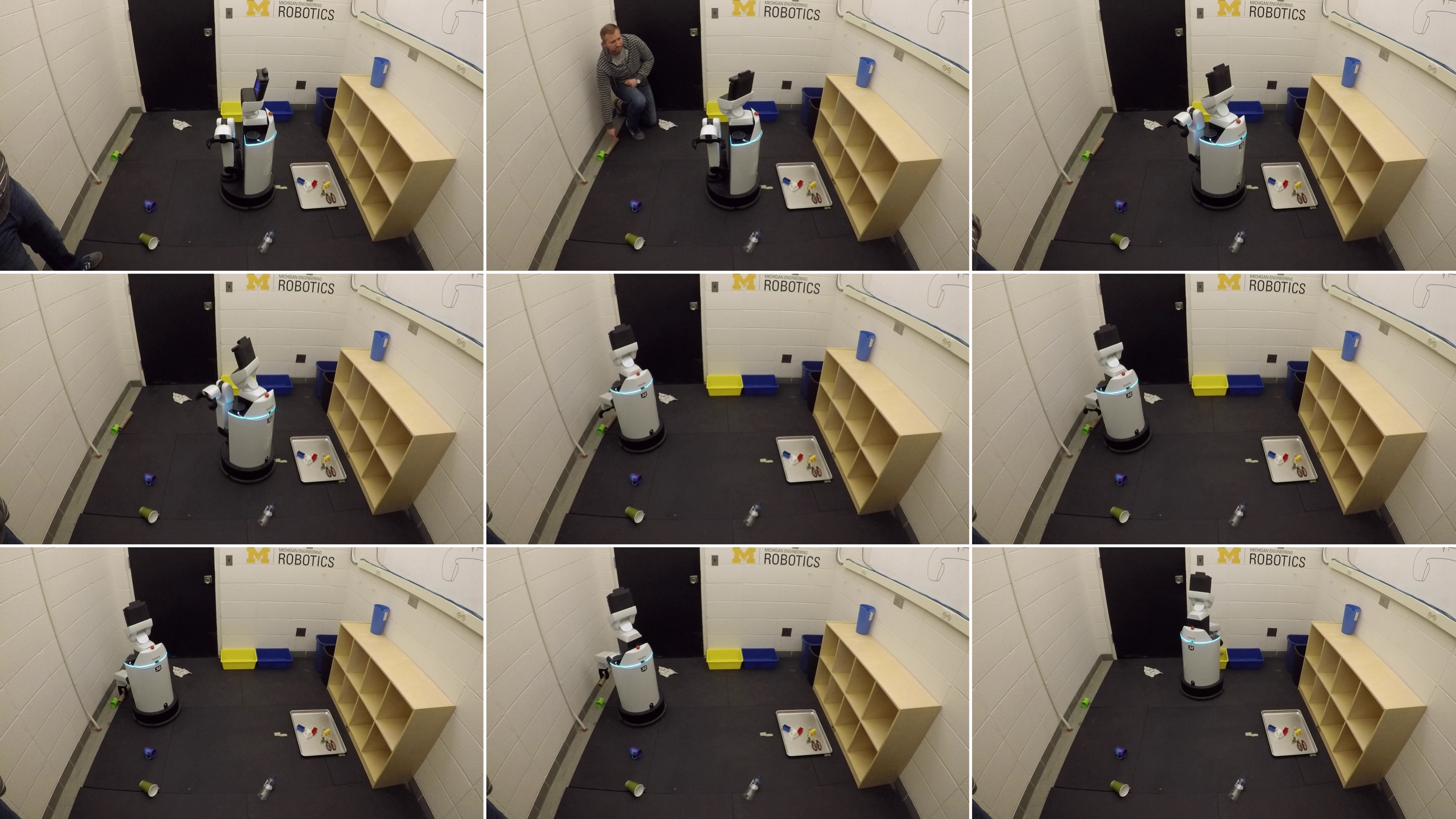}
	\caption{\textbf{Using VOS-VS for Human-Robot Collaboration}. HSR is asked to perform a pick-and-place task with the paper towel roll, but has no idea where it is. Using $\mathbf{H}_{\text{head}}$ VOS-VS, HSR tracks the person so that he can show HSR where to find the roll (top row).
		Using $\mathbf{H}_{\text{head}}$ VOS-VS again, HSR centers it's gaze on the roll to locate it, then uses $\mathbf{H}_{\text{base grasp}}$ to position itself for VOS-Grasp (middle row).
		Finally, HSR grasps the paper towel roll, verifies the grasp using our visual check, and then places the roll in the yellow bin (bottom row). 
		All pick-and-place experiment and trial videos are available at: \url{https://www.youtube.com/playlist?list=PLz52BAn_JPx8nVgP2XfnG_9TCJj0DwC5y}.
	}
	\label{fig:human}
\end{figure*}

\vspace{5pt}
\noindent\textbf{Figures for Pick-and-place Experiments} Figure~\ref{fig:banana} shows a fully-automated pick-and-place task. Figure~\ref{fig:human} shows a pick-and-place task with VOS-VS-based human collaboration.

{\small
\bibliographystyle{ieee}
\bibliography{vosvsbib}
}

\end{document}